\documentclass[nohyperref]{article}

\usepackage{microtype}
\usepackage{graphicx}
\usepackage{booktabs} 

\usepackage{hyperref}

\usepackage[accepted]{icml2023}

\usepackage{amsmath}
\usepackage{amssymb}
\usepackage{pifont}
\usepackage{mathtools}
\usepackage{amsthm}

\usepackage[capitalize,noabbrev]{cleveref}

\usepackage{listings}
\usepackage{colortbl}
\usepackage{xcolor}
\usepackage{float}
\usepackage{amsmath}
\usepackage{amssymb}
\usepackage{graphicx}
\usepackage{booktabs}
\usepackage{bbm}
\usepackage{bm}
\usepackage{soul}
\usepackage{caption}
\usepackage{subcaption}
\usepackage{multirow}
\usepackage{arydshln}
\usepackage[normalem]{ulem}
\usepackage{xspace}
\useunder{\uline}{\ul}{}

\newcommand{\patks}[1]{\textsc{pass}@$#1$\xspace}
\newcommand{\eg}{\textit{e.g., \xspace}}
\newcommand{\ie}{\textit{i.e., \xspace}}

\newcommand{\CodeLMp}{P_{\mathbf{LM}}}
\newcommand{\hy}{\hat{y}}
\newcommand{\hz}{\hat{z}}
\newcommand{\hs}{\hat{S}}
\newcommand{\cmark}{\ding{51}}
\newcommand{\xmark}{\ding{55}}
\newcommand{\bt}[1]{\textcolor{blue}{#1}}
\newcommand{\rt}[1]{\colorbox{pink}{#1}}

\newcommand{\ours}{\textsc{Lever}\xspace}
\newcommand{\oursf}{\textsc{Lever}\includegraphics[scale=0.10]{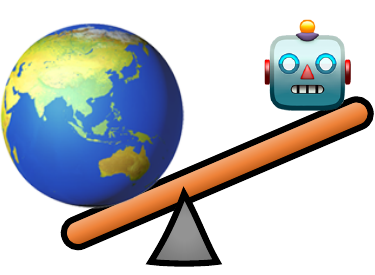}\xspace}
\newcommand{\model}[2]{#1-#2}

\newcommand{\hide}[1]{}

\iftrue
    \newcommand{\draftcomment}[3]{}
\else
    \newcommand{\draftcomment}[3]{\textcolor{#2}{{\bf\small [#1: #3]}}}
    
\fi
\newcommand{\victoria}[1]{\draftcomment{Victoria}{teal}{#1}}

\newcommand{\an}[1]{\draftcomment{AN}{cyan}{#1}}

\theoremstyle{plain}

\theoremstyle{definition}

\theoremstyle{remark}

\usepackage[textsize=tiny]{todonotes}

\icmltitlerunning{\ours: Learning to Verify Language-to-Code Generation with Execution}

\begin{document}

\twocolumn[
\icmltitle{\textsc{Lever}: 
Learning to Verify Language-to-Code Generation with Execution}



\icmlsetsymbol{equal}{*}
\icmlsetsymbol{intern}{$\dagger$}

\begin{icmlauthorlist}
\icmlauthor{Ansong Ni}{yale,intern}
\icmlauthor{Srini Iyer}{meta}
\icmlauthor{Dragomir Radev}{yale}
\icmlauthor{Ves Stoyanov}{meta}
\icmlauthor{Wen-tau Yih}{meta}
\icmlauthor{Sida I. Wang}{meta,equal}
\icmlauthor{Xi Victoria Lin}{meta,equal}
\end{icmlauthorlist}

\icmlaffiliation{yale}{Yale University}
\icmlaffiliation{meta}{Meta AI}

\icmlcorrespondingauthor{Ansong Ni}{ansong.ni@yale.edu}
\icmlcorrespondingauthor{Xi Victoria Lin}{victorialin@meta.com}
\icmlcorrespondingauthor{Sida I. Wang}{sida@meta.com}

\icmlkeywords{Machine Learning, ICML}

\vskip 0.3in
]



\printAffiliationsAndNotice{$^\dagger$Majority of the work done during an internship at Meta AI. \quad\icmlEqualContribution} 
\begin{abstract}
The advent of large language models trained on code (code LLMs) has led to significant progress in language-to-code generation. State-of-the-art approaches in this area combine LLM decoding with sample pruning and reranking using test cases or heuristics based on the execution results. 
However, it is challenging to obtain test cases for many real-world language-to-code applications, and heuristics cannot well capture the semantic features of the execution results, such as data type and value range, which often indicates the correctness of the program.
In this work, we propose \ours, a simple approach to improve language-to-code generation by learning to verify the generated programs with their execution results.
Specifically, we train verifiers to determine whether a program sampled from the LLMs is correct or not based on the natural language input, the program itself and its execution results. 
The sampled programs are reranked by combining the verification score 
with the LLM generation probability, and marginalizing over programs with the same execution results. On four datasets across the domains of table QA, math QA and basic Python programming, \ours consistently improves over the base code LLMs(4.6\% to 10.9\% with \texttt{code-davinci-002}) and achieves new state-of-the-art results on all of them.
\end{abstract}

\section{Introduction}
\begin{figure*}
    \centering
    \includegraphics[width=\linewidth]{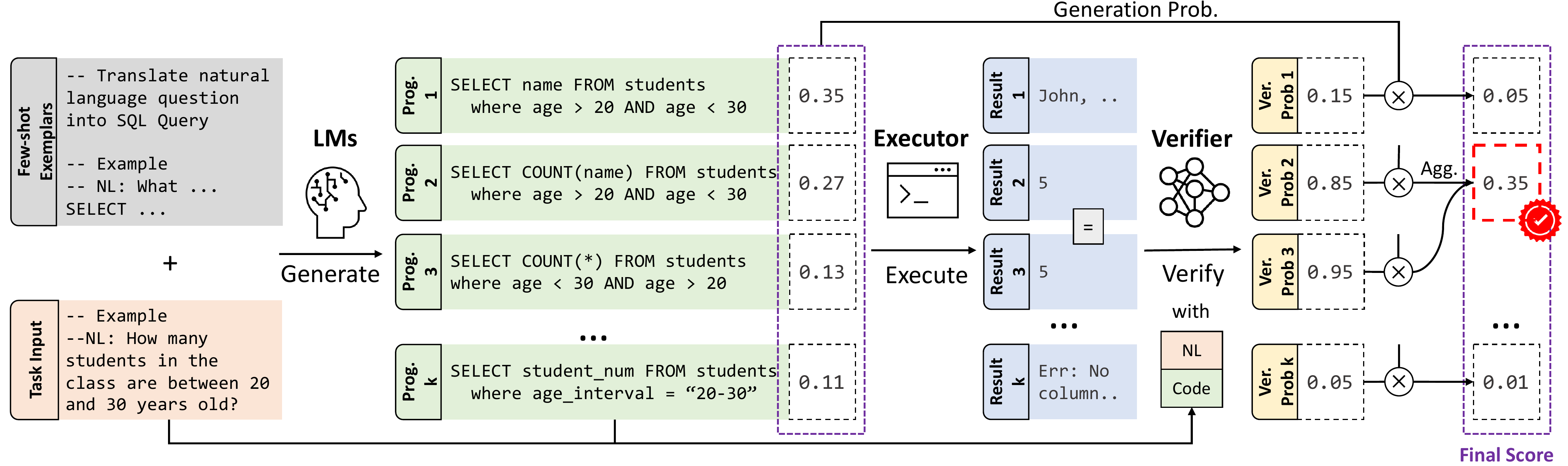}
    \caption{The illustration of \ours using text-to-SQL as an example. It consists of three steps: 1) \textit{Generation}: sample programs from code LLMs based on the task input and few-shot exemplars; 2) \textit{Execution}: obtain the execution results with program executors; 3) \textit{Verification}: using a learned verifier to output the probability of the program being correct based on the NL, program and execution results. 
    }
    \label{fig:pipeline}
\end{figure*}

The ability of mapping natural language to executable code is the cornerstone of a variety AI applications such as database interfaces~\citep{pasupat-liang-2015-compositional, yu-etal-2018-spider, shi-etal-2020-potential}, 
robotics control~\citep{zhou2021hierarchical,shridhar2020alfred} and virtual assistants~\citep{agashe-etal-2019-juice,lai2022ds}. Recent advances on large language models (LLMs)~\citep{brown2020language, wei2021finetuned, chowdhery2022palm}, especially those pre-trained on code (code LLMs)~\citep{chen2021codex,fried2022incoder,nijkamp2022codegen,li2022alphacode}, have shown great promise in such tasks with in-context few-shot learning~\cite{shi2022mbr,chen2022codet,zhang2022coderreviewer}. 
Yet their performance is still far from perfect~\cite{chen2021codex}.
Considering the computation cost to finetune such models, it is appealing to explore ways to improve them without changing their parameters. 

A key observation is that while LLMs struggles with precision in the few-shot setting, 
it often produces the correct output when enough samples are drawn. 
Previous work have shown that 
majority voting and filtering by test cases can significantly boost their performance when samples are drawn at scale~\cite{chen2021codex,austin2021mbpp,li2022alphacode}. 
~\citet{shen2021generateandrank} and~\citet{cobbe2021verifier} further demonstrated the effectiveness of training a verifier and using the verification scores to rerank the candidate solutions for math world problems.
Comparing to approaches that solely rely on execution consistency and error pruning, trained verifiers can make use of the rich semantic features in the model solutions, such as data types, value range, and variable attributes, which can be strong indicators of correctness of the programs.
While~\citet{cobbe2021verifier} and subsequent work~\cite{li2022diverse,kadavath2022knowwhattheyknow} focus on verifying natural language solutions by LMs, a natural question is whether the same approach can be applied to program solutions.

In this work, we propose \underline{le}arning to \underline{ver}ify (\oursf) language-to-code generation by code LLMs, with the help of execution.
More specifically, we train a verifier that learns to distinguish and reject incorrect programs based on the joint representation of the natural language description, the program surface form and its execution result. We further combine the verification probability with the LLM generation probability 
and 
marginalize over programs with the same execution results. 
We use this aggregated probability as the reranking score and output the programs that execute to the most probable result. 

We conduct extensive experiments on four different language-to-code benchmarks across domains of text-to-SQL semantic parsing, table QA, math reasoning and basic Python programming.
Experiment results with three different code LLMs show that \ours consistently improves the execution accuracy of the generated programs.
Notably, \ours coupled with \texttt{code-davinci-002} improves over strong baselines that use execution error pruning by 4.6\% to 10.9\%, and achieves the new state-of-the-art results on all four benchmarks, without using task-specific model architecture or prompting methods. 
Ablation studies show that execution results are crucial for the verification and \ours also yields non-trivial improvements in low-resource and weakly-supervised settings.\footnote{We open-source our experiment code for reproducibility: \url{https://github.com/niansong1996/lever}. }\footnote{Following~\citet{cobbe2021verifier}, by ``verifying'' we mean assessing whether a program's functionality matches its programmer's intent. This is distinct from the notion of formal verification in programming languages~\cite{10.5555/2930832}.}

\section{Approach}
 We now introduce the detailed formulation and training procedures of \ours. The key components are illustrated in \autoref{fig:pipeline}.
\subsection{Language-to-Code Generation with Code LLMs}
\label{sec:problem-definition}
The input for a language-to-code task typically consists of the natural language (NL) description and optionally some programming context (\eg data stores, assertions and so on). We denote such input as $x$. Given $x$,
a generation model $P(y|x)$ generates a program $y$ which is later executed via an executor $\mathcal{E}(\cdot)$ to obtain the result\footnote{Some datasets such as Spider~\cite{yu-etal-2018-spider} require the input values to be generated together with the programs, hence $y^*$ is directly executable. Others require the programs to be executable on separately provided test cases,~\eg MBPP~\cite{austin2021mbpp}. We adopt this notation for simplicity.
} $\mathcal{E}(y)$.
For few-shot learning with large LMs, the generation is also often conditioned on 
a fixed set of $m$ exemplars, $\{(x_i, y_i)\}_{i<m}$. 
Thus the few-shot language-to-code generation with code LLMs can be formulated as:
\begin{equation}
   \CodeLMp(y|x) = P(y|\operatorname{prompt}(x, \{(x_i, y_i)\}_{i<m})),
\end{equation}
where $\operatorname{prompt}(x, \{(x_i, y_i)\}_{i<m})$ is a string representation of the overall input.
Greedy search is typically used to find the program with the (approximately) highest generation probability, \ie $\hy_{\text{greedy}} \thickapprox \arg\max_{y}\CodeLMp(y|x)$.

\subsection{Reranking of Program Candidates}
\label{sec:reranking-method}
The key observation motivating our method is that a reasonably large sample set from $\CodeLMp(y|x)$ often includes the correct programs. 
This suggests that reranking of the program candidates may yield significant result improvement.
The idea of discriminative reranking~\citep{shen2004discriminative, collins2005discriminative} is to learn a scoring function $R(x, \hy)$ that measures how likely $\hy$ is the best output for input $x$. Given $R(\cdot)$, the reranker outputs the program with the highest reranking score among the set of candidates $S$:
\begin{equation}
\label{eq:rerank}
    \hy_\text{rerank} = \arg\max_{\hy\in S} R(x, \hy)
\end{equation}
Next we introduce how we adopt a trained verifier to verify and rerank program candidates sampled from code LLMs such that $\hy_\text{rerank}$ is better than $\hy_\text{greedy}$.
\paragraph{Program Sampling from Code LLMs.} 
Given input $x$, instead of performing greedy search, we obtain $k$ programs from $\CodeLMp(y|x)$ with temperature sampling, \ie $\{\hy_i\}_{i=1}^k\sim \CodeLMp(y|x)$. 
As the same programs may be sampled more than once, we perform deduplication to form a set of $n$ \textit{unique} program candidates $S=\{\hy_i\}_{i=1}^n$, where $n\leq k$.
We choose to do sampling instead of beam search mainly for two reasons: 1) recent work suggests that beam search for code generation typically results in worse performance due to degenerated programs \citep{austin2021mbpp, zhang2022coderreviewer}; and 2) beam search is not available or efficiently implemented for all LLMs that we test on (\eg Codex).

\paragraph{Verification with Execution.} 
We use a simple concatenation of the problem description $x$, candidate program $\hy$ and a representation of its execution results $\mathcal{E}(\hy)$ as the input to the reranker. 
Inspired by recent work \citep{cobbe2021verifier, li2022diverse}, we parameterize our discriminative reranker as a verification (\ie binary classification) model $P_{\theta}(v|x, \hy, \mathcal{E}(\hy))$, where $v\in\{0, 1\}$. In practice, the reranker can be implemented using any binary classification architecture. We report experiments using T5~\cite{colin2020t5} and RoBERTa~\cite{liu2019roberta} in \S\ref{sec:base-model-ablation}.

Given an input $x$ and a candidate program $\hy\in S$, we obtain the reranking probability as the joint probability of generation and passing the verification:
\begin{equation}
\label{eq:join-reranking-prob}
    P_R(\hy, v_{=1}|x) = \CodeLMp(\hy|x) \cdot P_{\theta}(v_{=1}|x, \hy, \mathcal{E}(\hy))
\end{equation}
\paragraph{Execution Result Aggregation.} Since programs with the same semantics may have different surface forms, we further aggregate the reranking probability of the programs in $S$ that executes to the same result. In this way, we relax the dependency on the surface form and focus on the execution results instead. The final scoring function for reranking is therefore:
\begin{equation*}
    R(x, \hy) = P_R(\mathcal{E}(\hy), v_{=1}|x) = \sum_{y\in S, \mathcal{E}(y)=\mathcal{E}(\hy)} P_R(y, v_{=1}|x)
\end{equation*}
Since there might be several programs that share the same execution result of the highest probability, we break tie randomly in this case when outputting the programs.

\subsection{Learning the Verifiers}
\label{sec:learning-reranker}
The previous sections described how to use a verifier 
at inference time. 
Next we introduce its training process. 
\paragraph{Training Data Creation.} For language-to-code datasets, each example is typically a triplet of $(x, y^*, z^*)$, where $z^*=\mathcal{E}(y^*)$ is the gold execution result and $y^*$ is the gold program.
As annotating the programs requires domain expertise, for some datasets where the final results can be directly obtained, only $z^*$ but no $y^*$ is provided for learning \cite{artzi2013weakly, cheng2018weakly, goldman2018weakly}. This is known as the weakly-supervised setting. 
To gather training data, we obtain a set of $n$ unique programs candidates $\hs=\{\hy_i\}_{i=1}^n$ for each input $x$ in the training set, by first sampling $k$ programs from $\CodeLMp(\hy | x)$ and then remove all the duplicated programs, similarly as inference time. 
Then for each program candidate $\hy\in S$, we obtain its binary verification label by comparing the execution result $\hz=\mathcal{E}(\hy)$ with the gold\footnote{For datasets that provide multiple test cases, we label a program as correct if and only if its execution results match the ground truth program on all test cases.} execution result $z^*$, \ie $v=\mathbbm{1}(\hz=z^*)$.
For the datasets that contain the gold program $y^*$, we append $(x, y^*, z^*, v_{=1})$ as an additional verification training example, and we skip this step for the weakly-supervised datasets.
This way, we create a set of verification training examples $\{(x, \hy_i, \hz_i, v_i) \mid \hy_i\in S\}$ for each input $x$. 
\paragraph{Learning Objective.} Given this set of verification training examples, we formulate the loss for input $x$ with the negative log-likelihood function, normalized by the number of program candidates
\begin{equation}
\label{eq:obj}
    \mathcal{L}_\theta(x, S) = -\frac{1}{|S|}\cdot\sum_{\hy_i\in S}\log P_{\theta}(v_i | x, \hy_i,\hz_i)
\end{equation}
The normalization step is important to prevent an example with a large number of unique program candidates to dominate learning.

\section{Experimental Setup}
\begin{table}[t!]
\small
\centering
    \begin{tabular}{p{0.2\linewidth}p{0.14\linewidth}p{0.14\linewidth}p{0.14\linewidth}p{0.13\linewidth}}
    \toprule
                        & \textbf{Spider}     & \textbf{WikiTQ}   & \textbf{GSM8k}    & \textbf{MBPP}\\\midrule
    Domain              & Table QA  & Table QA & Math QA & Basic Coding \\
    Has program & \cmark & \cmark$^*$ & \xmark & \cmark \\
    Target & SQL & SQL  & Python & Python \\ 
    \midrule
    \multicolumn{5}{c}{\textit{Data Statistics}} \\\midrule
    \# Train             & 7,000    & 11,321      & 5,968       & 378              \\
    \# Dev               & 1,032    & \;\;2,831       & 1,448   & \;\;90         \\
    \# Test              & \;\;-    & \;\;4,336       & 1,312   & 500  \\
    \midrule
    \multicolumn{5}{c}{\textit{Few-shot Generation Settings}} \\\midrule
    Input \quad Format & Schema + NL & Schema + NL & NL & Assertion + NL \\
    \# Shots & 8$^\ddag$ & 8 & 8 & 3 \\
    \# Samples & \multirow{2}{*}{20/50$^\dag$} & \multirow{2}{*}{50/50}  & \multirow{2}{*}{50/100} & \multirow{2}{*}{100/100} \\
    (train / test)\\
    Generation & \multirow{2}{*}{128} & \multirow{2}{*}{128}  & \multirow{2}{*}{256} & \multirow{2}{*}{256} \\
    Length\\
    \bottomrule
    \end{tabular}
    \caption{Summary of the datasets used in this work. $^*$: About 80\% examples in WikiTableQuestions are annotated with SQL by \citet{shi-etal-2020-potential}. 
    $^\dag$: 50/100 for InCoder and CodeGen for improving the upper-bound. 
    $^\ddag$: Only the first 2 of the 8 exemplars are used for InCoder and CodeGen due to limits of context length and hardware.
    }
    \label{tab:datasets}
\end{table}
\subsection{Datasets}
We conduct experiments on four language-to-code datasets across domains of semantic parsing, table QA, math reasoning and basic python programming.
The main settings of these four datasets are shown in \autoref{tab:datasets}. More detailed settings for verification are in \autoref{tab:hyperparams} of the Appendix.

\noindent$\triangleright$ \textbf{Spider} \citep{yu-etal-2018-spider} is a semantic parsing dataset on generating SQL queries from natural language questions. With 7k parallel training data, it is also ideal for finetuning generators; \\
\noindent$\triangleright$ \textbf{WikiTableQuestions (WikiTQ)} \citep{pasupat-liang-2015-compositional} is a table question answering dataset, for which we attempt to solve by generating and executing SQL queries over the source tables. We use the preprocessed tables from \citet{shi-etal-2020-potential} and adopt their annotated SQL queries for adding gold programs for the originally weakly-supervised dataset; \\
\noindent$\triangleright$ \textbf{GSM8k} \citep{cobbe2021verifier} is a benchmark for solving grade-school level math word problems. Following previous work \citep{chowdhery2022palm, chen2022program, gao2022pal}, we approach this benchmark by generating Python programs from questions in NL, which should produce the correct answer upon execution.
The original dataset only has natural language and not program solutions, thus it is weakly-supervised for language-to-code; \\
\noindent$\triangleright$ \textbf{MBPP} \citep{austin2021mbpp} contains basic Python programming programs stated in natural language. Each example is equipped with 3 test cases to check the correctness of the programs. Following previous work \citep{shi2022mbr, zhang2022coderreviewer}, we use the first test case as part of the prompt for the model to generate correct function signatures and use all three of them for evaluating correctness.

\subsection{Code LLMs}
We evaluate \ours with three different code LLMs: \\
\noindent$\triangleright$ \textbf{Codex} \citep{chen2021codex} is a family of code LLMs of different sizes developed by OpenAI. Specifically, we use the \texttt{code-davinci-002} API\footnote{\url{https://openai.com/api/}} through its official Python bindings. \\
\noindent$\triangleright$ \textbf{InCoder} \citep{fried2022incoder} is a family of code LLMs up to 6B parameters trained on a large corpus of code with  permissively licenses.
We experiment with \model{InCoder}{6B} and use it for left-to-right generation. \\
\noindent$\triangleright$ \textbf{CodeGen} \citep{nijkamp2022codegen} is a family of code LLMs and we evaluate the \model{CodeGen}{16B-multi} version. 
Although SQL files are not included in the training corpus for CodeGen,
we found it to still perform reasonably well on SQL generation tasks possibly because the SQL queries were mixed in with source files of other programming languages.

\subsection{Baselines and Evaluation Metric}
\label{sec:metric-baselines}
\paragraph{Baselines.}
\label{sec:baseline-details}
We compare~\ours to the following baseline approaches for generating programs using code LLMs. \\
\noindent$\triangleright$ \underline{\textbf{Greedy}}: Select the most likely token per decoding step. \\
\noindent$\triangleright$ \underline{\textbf{Maximum Likelihood (ML)}}: 
From $k$ sampled program candidates, select the program with the highest generation log-probability, \ie $\log\CodeLMp(\hy|x)$ (or normalized generation log-probability as $\log\CodeLMp(\hy|x)/|\hy|$). We determine empirically using the development set whether to use the normalized probability for each dataset. More details can be found in \autoref{sec:more-imp-details}. \\
\noindent$\triangleright$ \underline{\textbf{Error Pruning + ML (EP + ML)}}: Prune out the candidate programs with execution errors; then select the program with the maximum likelihood; \\
\noindent$\triangleright$ \underline{\textbf{Error Pruning + Voting (EP + Voting)}}:
Take the majority vote on the execution results among the error-free programs, and select the most-voted execution result and its corresponding programs.

We focus on comparing with the EP+ML baseline, as it is a simple reranking method that exploits execution and yields competitive results consistently across different datasets and code LLMs.

\paragraph{Evaluation metric.}
Following previous work~\citep{xie2022unifiedskg, liu2021tapex, ni2022learning, zhang2022coderreviewer}, we use \textit{execution accuracy} as the main evaluation metric for all datasets, which measures the percentage of examples that yields the gold execution result or pass all test cases. 

\subsection{Implementation Details} 

\paragraph{Verifier training.} 
\an{I feel this paragraph is not that important, we should move it to the appendix. Note that how the verification training data is created is already explained in section 2.2}\victoria{For arXiv there is no space pressure and reproducibility is more important. Let's keep this one here and move to appendix for camera ready.}
\an{I know but I feel like this belongs to the additional implementation details, as the ones we describe in \autoref{sec:more-imp-details}. And the settings are better explained when the text is closer with \autoref{tab:hyperparams}, as it is referenced multiple times here.}
We create the verification training data by sampling from the LLMs on the training set, using the sampling budget described in \autoref{tab:datasets}. More statistics of the resulting training data can be found in \autoref{tab:hyperparams} in the Appendix. 
When learning the verifiers, as shown in \autoref{eq:obj}, the training loss is computed by averaging over all the program samples for each example. 
As we batch the program samples for the same examples together, the effective batch size will also be multiplied by the sample size.
\victoria{I'll leave this part to you to finish, Ansong. I think it is better to describe it in two sentences. One explains your batching strategy and the second sentence explains the memory issue. Regarding batching strategy, what I don't understand is what do you mean by ``batch the program samples for the same examples together'', can the program samples for the same example spread across multiple batches?}\an{yes, it can. but we used to do contrastive learning (rankin-based loss) which requires the sampled programs in the same batch so they can contrast each other. thus our implementation always batches the programs from the same example together}
This could be problematic when sample size gets large (up to 100 in our experiments) as they may not be able to fit into the GPU memory at once. Therefore, we down-sample the programs used for learning per example in each iteration.
The random down-sampling happens at the beginning of every epoch of training so the verifiers are able to see different programs each epoch.
Detailed batch sizes and downsampling factor can be found in \autoref{tab:hyperparams} in the Appendix.

\paragraph{Execution result representation.}
The input to the verifier is a concatenation of the task input, the candidate program and its execution results. For Spider and WikiTQ, we use the linearized resulting tables from SQL execution as the execution results. For GSM8k, we use the value of the variable named ``\texttt{answer}'' after executing the program as the execution results. For MBPP, we use the type and value (casted to string) returned by the functions. All execution errors are represented as ``\texttt{ERROR: [reason]}'', such as ``\texttt{ERROR: Time out}''.
Examples of these verifier inputs for different datasets can be found in \autoref{tab:verifier-inputs}.

\paragraph{Verifier model selection.}
We use the development set to choose the best verifier model.
We select T5-base for Spider, T5-large for WikiTQ and MBPP, and RoBERTa-large for GSM8k as the base LM for the verifiers to use in the main experiments\footnote{We attempted using the code LLM itself as the verifier in a few-shot manner, but the performance is inferior than EP+ML.
}. 
The selection process is detailed in \autoref{sec:base-model-ablation}.
For the T5 models~\cite{colin2020t5}, we train them to output the token ``yes/no'' for each positive/negative example given the verifier input, and we take the probability of generating ``yes'' as the verification probability during inference. For RoBERTa~\cite{liu2019roberta}, we add a linear layer on top of the \texttt{[CLS]} head, following the standard practice of sequence classification with encoder-only models~\cite{devlin2019bert}.

The details of LLM sampling, few-shot prompt construction and dataset-specific setups can be found in \autoref{sec:more-imp-details}.   

\begin{table}[]
    \centering
    \small
    \begin{tabular}{ll}
        \toprule
        \textbf{Methods}            & \textbf{Dev}                    \\\midrule
        \multicolumn{2}{c}{\textit{Previous Work without Finetuning}} \\
        \citet{rajkumar2022evaluating}   & 67.0      \\
        MBR-Exec \citep{shi2022mbr} & 75.2 \\
        Coder-Reviewer \citep{zhang2022coderreviewer} & 74.5 \\\midrule
        \multicolumn{2}{c}{\textit{Previous Work with Finetuning}} \\
        T5-3B \citep{xie2022unifiedskg}  & 71.8      \\
        PICARD \citep{scholak2021picard} & 75.5      \\
        RASAT \citep{qi2022rasat} & 80.5             \\\midrule
        \multicolumn{2}{c}{\textit{This Work with code-davinci-002}} \\
        Greedy                        & 75.3                    \\
        EP + ML                       & 77.3 \\
        \oursf                        & \textbf{81.9}$_{\pm0.1}$            \\
        \bottomrule
    \end{tabular}
    \caption{Execution accuracy on the Spider dataset. Standard deviation is calculated over three runs with different random seeds (same for the following tables when std is presented).
    }
    \vspace{-1pt}
    \label{tab:codex-spider}
\end{table}
\begin{table}[]
    \centering
    \small
    \begin{tabular}{lll}
        \toprule
        \textbf{Methods}     & \textbf{Dev}      & \textbf{Test}        \\\midrule
        \multicolumn{3}{c}{\textit{Previous Work without Finetuning}} \\
        Codex QA$^*$ \citep{cheng2022binding}  & 50.5 & 48.7  \\
        Codex SQL \citep{cheng2022binding}     & 60.2 & 61.1  \\
        Codex Binder \citep{cheng2022binding}  & \textbf{65.0} & 64.6  \\\midrule
        \multicolumn{3}{c}{\textit{Previous Work with Finetuning}} \\
        TaPEX$^*$ \citep{liu2021tapex}       & 60.4 & 59.1  \\
        TaCube \citep{zhou2022tacube}    & 61.1 & 61.3  \\
        OmniTab$^*$ \citep{jiang2022omnitab} & -    & 63.3  \\\midrule
        \multicolumn{3}{c}{\textit{This Work with code-davinci-002}} \\
        Greedy                 & 49.6             & 53.0          \\
        EP + ML                & 52.7             & 54.9          \\
        \oursf                 & 64.6$_{\pm0.2}$  & \textbf{65.8}$_{\pm0.2}$    \\
        \bottomrule
    \end{tabular}
    \caption{Execution accuracy on the WikiTQ dataset. $^*$: modeled as end-to-end QA without generating programs as a medium. 
    }
    \vspace{-1pt}
    \label{tab:codex-wtq}
\end{table}

\section{Main Results}
We show the performance of \ours coupled with Codex-Davinci and compare it with the state-of-the-art finetuning and few-shot performances from previous work for Spider (\autoref{tab:codex-spider}), WikiTQ (\autoref{tab:codex-wtq}), GSM8k (\autoref{tab:codex-gsm}) and MBPP (\autoref{tab:codex-mbpp}). 
In addition, we also evaluate \ours with InCoder and CodeGen models on Spider and GSM8k (\autoref{tab:codelm-ablations}).

\subsection{Effectiveness of \ours.} 
\ours consistently improves the performance of all code LLMs on all tasks, yielding improvements of 6.6\% (Spider) to 17.3\% (WikiTQ) over the greedy decoding baselines for \model{Codex}{Davinci}. For weaker models such as InCoder and CodeGen, we observe improvements up to 30.0\% for Spider and 15.0\% for GSM8k. Moreover, 
\ours combined with Codex-Davinci also achieves new state-of-the-art results on all four datasets, with improvements ranging from 1.2\% (WikiTQ) to 2.0\% (MBPP). 
On the challenging text-to-SQL dataset, Spider, where the previous state-of-the-art is achieved by finetuning a T5-3B model augmented with relational-aware self-attention, 
we achieved even better results with Codex-Davinci + \ours, where the verifier is finetuned using a T5-base model.
\ours also improves the previous best results on Spider using InCoder and CodeGen, by 13.2\% and 20.6\%, respectively.
\begin{table}[t]
    \centering
    \small
    \begin{tabular}{lll}
        \toprule
        \textbf{Methods}     & \textbf{Dev}      & \textbf{Test}    \\\midrule
        \multicolumn{3}{c}{\textit{Previous Work without Finetuning}} \\
        PAL \citep{gao2022pal} & - & 72.0 \\
        Codex + SC$^\dag$ \citep{wang2022self} & - & 78.0 \\
        PoT-SC \citep{chen2022program} & - & 80.0 \\\midrule
        \multicolumn{3}{c}{\textit{Previous Work with Finetuning}} \\
        Neo-2.7B + SS \citep{ni2022learning}       & 20.7 & 19.5 \\
        Neo-1.3B + SC \citep{welleck2022generating} & - & 24.2 \\
        DiVeRSe$^*$$^\dag$ \citep{li2022diverse} & - & 83.2 \\\midrule
        \multicolumn{3}{c}{\textit{This Work with codex-davinci-002}} \\
        Greedy                      & 68.1     & 67.2    \\
        EP + ML                     & 72.1     & 72.6    \\
        \oursf                      & \textbf{84.1}$_{\pm0.2}$    & \textbf{84.5}$_{\pm0.3}$    \\
        \bottomrule
    \end{tabular}
    \caption{Execution accuracy on the GSM8k dataset. $^*$: finetuned model combined with Codex
    (similar to \ours); $^\dag$: generating natural language solutions instead of programs.}
    \label{tab:codex-gsm}
\end{table}
\begin{table}[]
    \centering
    \small
    \begin{tabular}{lll}
        \toprule
        \textbf{Methods}     & \textbf{Dev}      & \textbf{Test}    \\\midrule
        \multicolumn{3}{c}{\textit{Previous Work without Finetuning}} \\
        MBR-Exec \citep{shi2022mbr}       & -       & 63.0 \\
        Reviewer \citep{zhang2022coderreviewer}       & -       & 66.9 \\\midrule
        \multicolumn{3}{c}{\textit{This Work with codex-davinci-002}} \\
        Greedy                                & 61.1    & 62.0    \\
        EP + ML                               & 62.2    & 60.2    \\
        \oursf       & \textbf{75.4}$_{\pm0.7}$    & \textbf{68.9}$_{\pm0.4}$    \\
        \bottomrule
    \end{tabular}
    \caption{Execution accuracy on the MBPP dataset. 
    }
    \label{tab:codex-mbpp}
\end{table}
\begin{figure*}[ht]
    \centering\includegraphics[width=\linewidth]{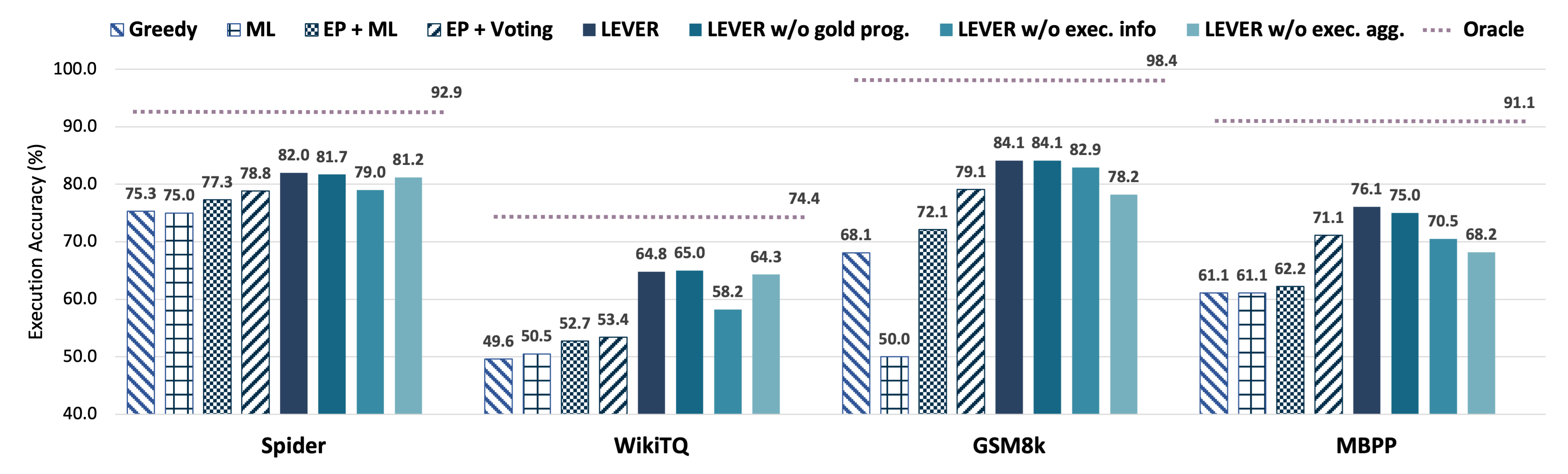}
    \caption{Comparison of \oursf and baselines with \model{Codex}{Davinci}. \ours and its ablation results are in solid bars.}
    \label{fig:codex-ablations}
\end{figure*}

As \ours is a simple method that combines few-shot LM generation with learned verifiers, it can potentially benefit more advanced prompting methods \citep{li2022diverse, cheng2022binding} or model architectures \citep{qi2022rasat, wang2020rat}, which we leave as future work.

\subsection{Ablations with \ours}
\label{sec:main-ablation}

We perform ablation study for \ours with \model{Codex}{Davinci} and compare with the baselines mentioned in \autoref{sec:baseline-details}, and the results are shown in \autoref{fig:codex-ablations}. The same ablations are conducted for InCoder and CodeGen with results in \autoref{tab:codelm-ablations}. In these results, we include an ``\underline{\textbf{Oracle}}'' performance which is obtained by always selecting the correct program as long as they appear in the sample set.

\begin{table}[t]
\footnotesize
\centering

\begin{tabular}{lccccc}
\toprule
\textbf{Methods}     & \multicolumn{2}{c}{\textbf{InCoder-6B}} &  & \multicolumn{2}{c}{\textbf{CodeGen-16B}} \\ \cline{2-3} \cline{5-6} 
            & \textbf{Spider}        & \textbf{GSM8k}         &  & \textbf{Spider}        & \textbf{GSM8k}         \\\midrule
\textit{Previous work:} \\
\quad \textsc{MBR-Exec} & 38.2 & - & & 30.6 & - \\
\quad Reviewer & 41.5 & - & & 31.7 & - \\\midrule
\textit{Baselines}: \\
\quad Greedy        & 24.1          & 3.1         &  & 24.6          & 7.1         \\
\quad ML            & 33.7          & 3.8         &  & 31.2          & 9.6         \\
\quad EP + ML       & 41.2          & 4.4         &  & 37.7          & 11.4        \\
\quad EP + Voting        & 37.4          & 5.9         &  & 37.1          & 14.2        \\\midrule
\oursf        & 54.1          & \textbf{11.9}        &  & 51.0          & \textbf{22.1}        \\
\quad $-$ gold prog. & 53.4          &      -      &  & \textbf{52.3}          &      -       \\
\quad $-$ exec. info & 48.5          & 5.6         &  & 43.0          & 13.4        \\
\quad $-$ exec. agg. & \textbf{54.7}          & 10.6        &  & 51.6          & 18.3        \\\midrule
Oracle             & 71.6          & 48.0        &  & 68.6          & 61.4        \\\bottomrule
\end{tabular}

\caption{Results with InCoder and CodeGen as the Code LLMs, evaluated on the dev set with T5-base as the verifier. Previous work results were copied from \citet{zhang2022coderreviewer}.
}
\label{tab:codelm-ablations}
\end{table}
\paragraph{Effect of including execution results.} 
\label{sec:exec-info-effectiveness}
According to \autoref{fig:codex-ablations}, the performance drops considerably on all four benchmarks when execution result is removed from the verifier input, indicating that the execution outcome is important for verifier training. 
The effect varies across different datasets. While it causes an absolute performance drop of 6.6\% and 5.6\% for WikiTQ and MBPP, as the drop is smaller for Spider (3.0\%) and GSM8k (1.2\%). 
We found the code samples for WikiTQ and MBPP contain more execution errors, which explains why our approach is more effective on these two datasets.
Table~\ref{tab:codelm-ablations} shows similar trends for InCoder-6B and CodeGen-16B on Spider and GSM8k. The smaller LMs have worse few-shot performance and removing the execution information from the verifier often results in even greater performance drops. 
\hide{
Moreover, we also observe that the effect of including execution results varies for different code LLMs and datasets.
This suggests that the verifiers effectively make use of the execution information to make decisions in these more difficult settings.
}
Moreover, we found that \ours in general outperforms the EP+ML baseline, indicating that the verifiers can make use of clues 
beyond simple execution errors.
More detailed quantitative analysis of when execution information helps is in \autoref{fig:quantitative-analysis}.

\paragraph{Effect of execution result aggregation.} Aggregating the programs with the same execution result is a simple and widely used technique \citep{chen2022program, cheng2022binding}. We find execution aggregation work well with \ours on datasets with Python output, but only marginally benefit the SQL datasets. 
A probable reason is that the Python code structure is more flexible than that of the domain-specific languages as SQL. In the database querying domain, it is more likely for an incorrect program to execute to some trivial but wrong results (\eg ``0'' or empty table). After aggregation, such incorrect results may accumulate enough probability mass to out-weight the correct one, leading to negative impact on the performance.

\paragraph{Weakly-supervised settings.} We also compare the performance of \ours under fully- and weakly-supervised settings. \autoref{fig:codex-ablations} and \autoref{tab:codelm-ablations} show that the performance of \ours is largely preserved when the gold programs are not given and the weakly-supervised setting is used (\S\ref{sec:learning-reranker}), with an absolute performance drop up to 1.1\%. This suggests that \ours works well under the weakly-supervised settings, and the program itself is less informative for verification comparing to the execution results. 

\section{Analysis}
\subsection{Training Example Scaling}
\label{sec:data-scaling}

\begin{figure}[t]
        \centering
        \includegraphics[width=0.95\linewidth]{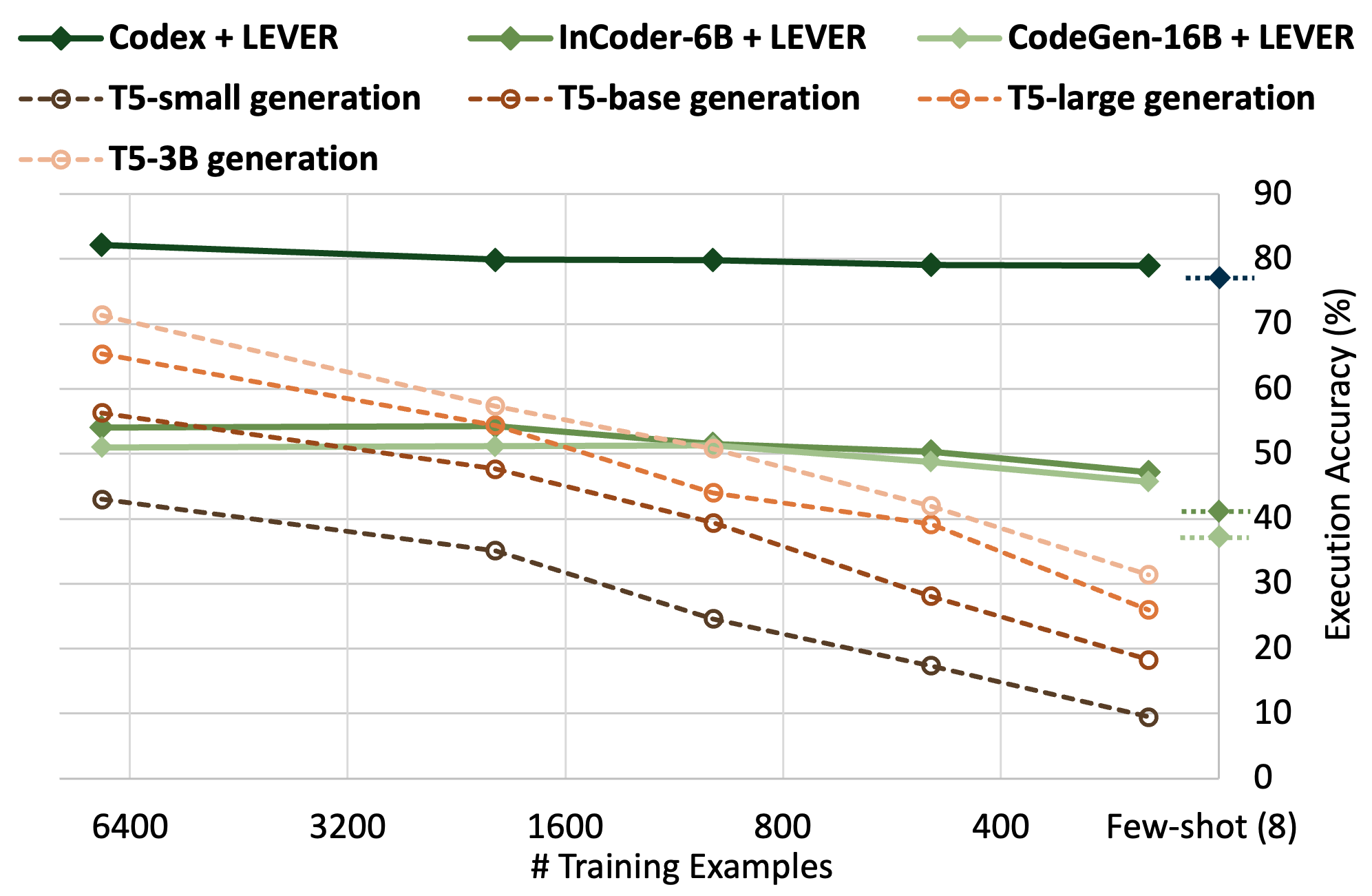}
    \caption{Verification vs. generation performance when decreasing the number of training examples for Spider. Data markers on the $y$-axis denote the EP+ML baseline, and the $x$-axis is on the logarithmic scale. 
    T5-base is used as the base model for \ours.
    WikiTQ and GSM8k results can be found in \autoref{fig:train-examples-ablation-wtq-gsm} in the Appendix. 
    }
    \label{fig:train-examples-ablation-spider}
\end{figure}
\begin{figure}[t]
    \centering
    \begin{subfigure}[b]{0.49\textwidth}
         \centering
         \includegraphics[width=\linewidth]{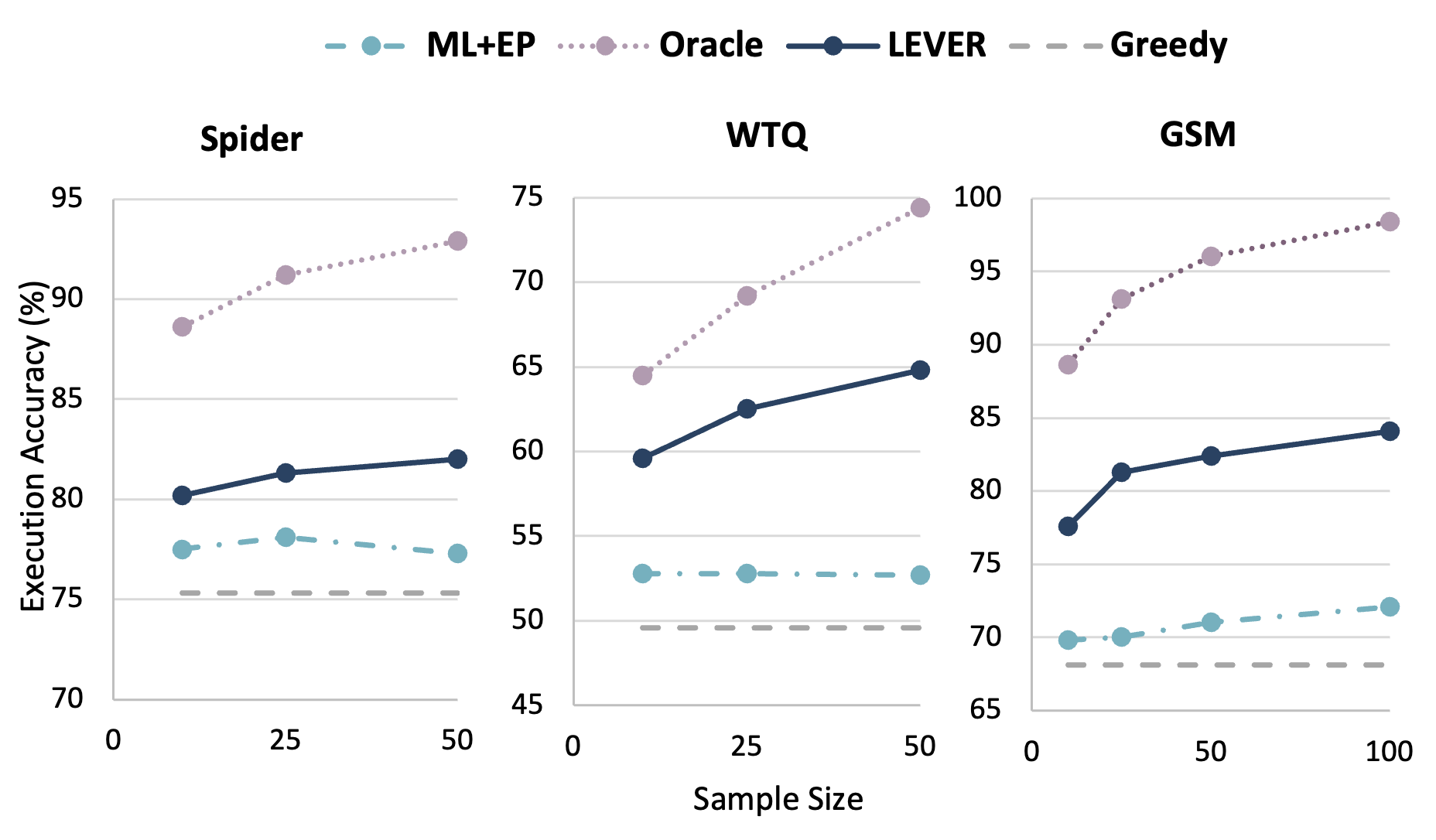}
         \caption{Ablation on sample size at inference time for \ours, while sample size at training time is fixed as in \autoref{tab:datasets}.}
         \label{fig:inf-sample-size-ablation}
    \end{subfigure}
    \newline
    \vspace{1pt}
    \newline
    \begin{subfigure}[b]{0.49\textwidth}
         \centering
         \includegraphics[width=\linewidth]{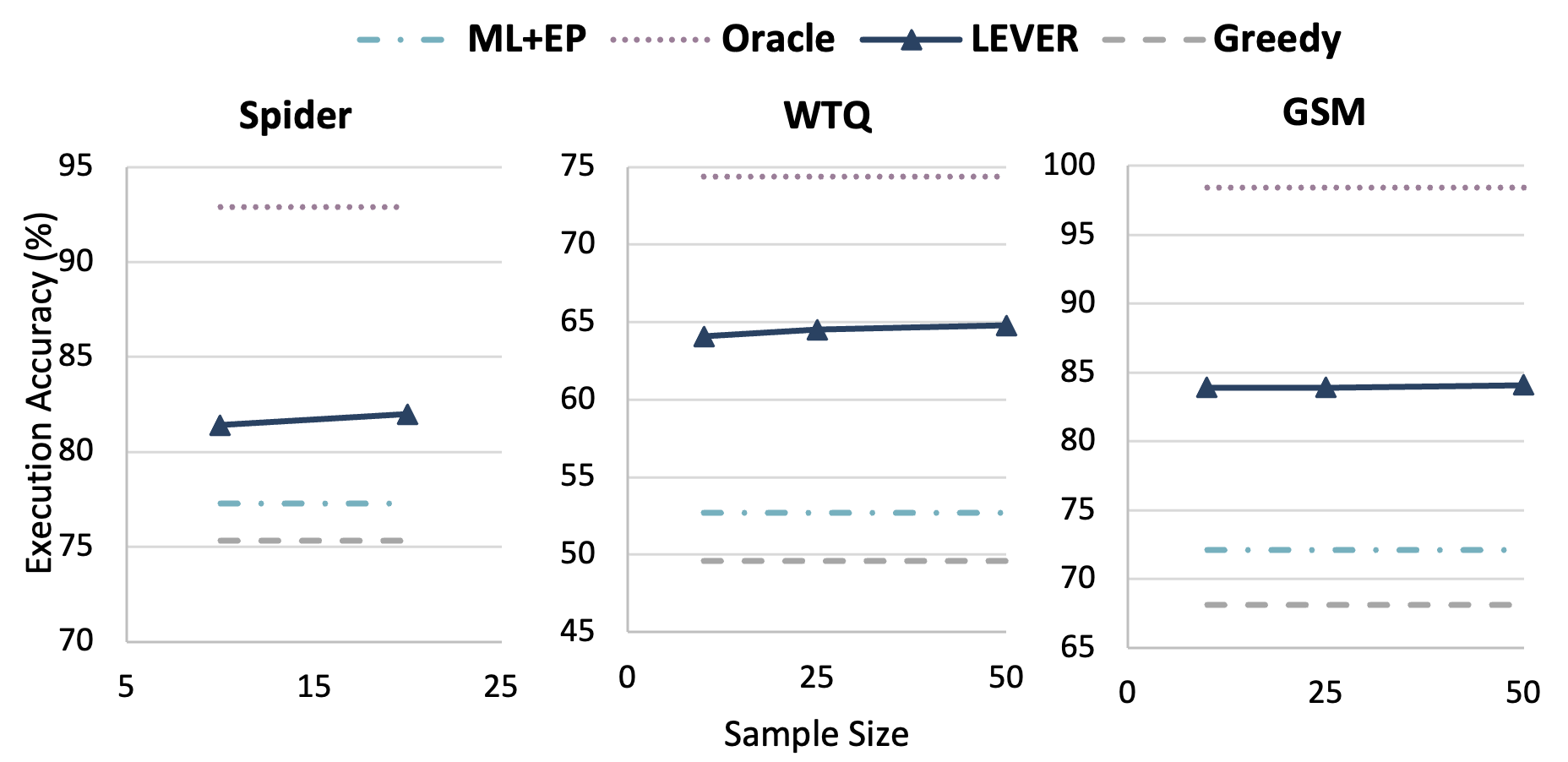}
         \caption{Performance with different number of programs to sample per example for training the verifiers. Sample size at inference time is fixed as in \autoref{tab:datasets}.} 
         \label{fig:train-sample-size-ablation}
     \end{subfigure}
     \caption{How sample size during training and inference time affects the performance, with \ours + Codex-Davinci.}
     \label{fig:sample-size-ablation}
\end{figure}
We show how the performance of \ours changes with fewer training examples in \autoref{fig:train-examples-ablation-spider}, using Spider as an example. More results on WikiTQ and GSM8k are in \autoref{sec:train-examples-ablation-wtq-gsm}.
The improvements with \ours over base LLMs are still consistent even when only 250 examples are given, with improvements ranging from 1.7\% to 10.0\% over different datasets and LLMs. This suggests that \ours can work under few-resource settings. 
Moreover, the trend also varies for different datasets and code LLMs, for example, 
 when using Codex as the LLM, the performance of \ours drops by 6.4\% for WikiTQ and only 3.2\% for Spider. However, also on Spider, the performance is lowered by 6.9\% and 5.3\% for InCoder and CodeGen. This suggests that having more training examples for \ours has larger effect for harder datasets and weaker LMs.

With \autoref{fig:train-examples-ablation-spider}, we also compare the performance of \ours with the T5 models being directly finetuned for generation given the same number of training examples. While verification can be learned with only hundreds of examples, the performance of finetuned T5 models drastically drops when less training examples are available. 
As an example, for 500 examples, a T5-base verifier on InCoder/CodeGen outperforms a finetuned T5-3B generator by $\sim7\%$.

\subsection{Sample Size Scaling} 
\label{sec:sample-size-analysis}
Since drawing samples from LLMs in may be costly computational-wise, here we study the how sample size during training and inference time affects the performance.
As we can see from \autoref{fig:inf-sample-size-ablation}, during inference time, when lowering the sample size from 50 to 10 programs per example, the performance of \ours drops by 1.8\% (Spider) to 5.2\% (WikiTQ). This indicates that the \ours is sensitive to the sample size at inference time, which is expected as it also greatly affects oracle results (\ie the upper-bound for reranking). In comparison, \autoref{fig:train-sample-size-ablation} shows that \ours is highly insensitive to the sample size for providing training data, with the performance gap all below 1\% for the three datasets. 
Overall, the results show that a higher sampling budget helps more at test time. 
\subsection{Verifier and Generator Calibration}
\begin{figure}
    \centering
    \includegraphics[width=\linewidth]{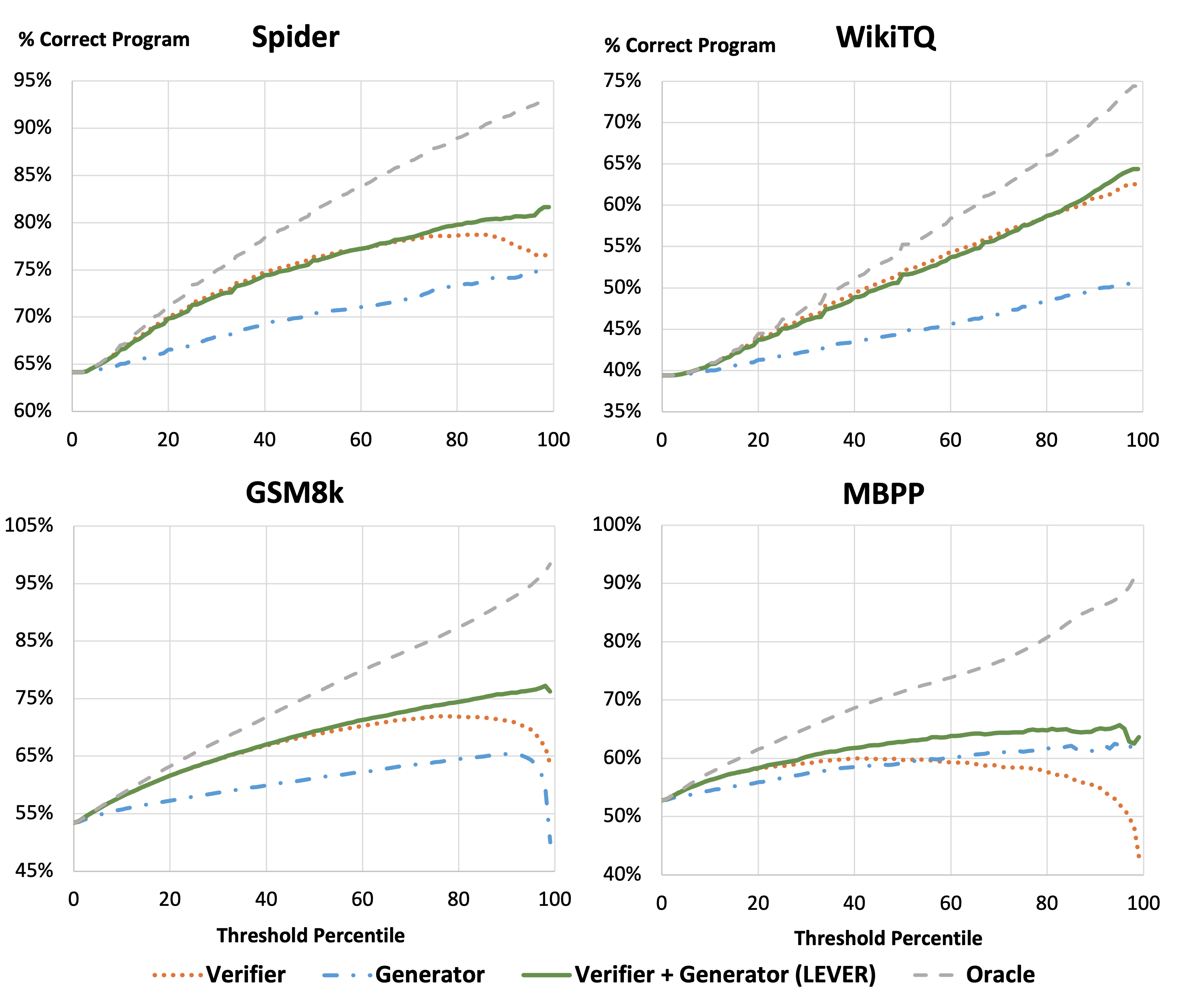}
    \caption{Calibration of the verifier, generator (Codex-Davinci), and their combined probability (used by \ours). The sampled programs are first ranked by the model probabilities. The $x$-axis represents the percentage of samples excluded after thresholding, and the $y$-axis represents the percentage of correct programs in the remaining samples. 
    Execution aggregation is not applied in this group of plots to ensure the scoring of different programs are independent.\an{deduplication is still applied, because it is percentile, it doesn't need to be 100 programs}}
    \label{fig:percentile_analysis}
\end{figure}
We study how well-calibrated are the verifier and generator in identifying correct programs. \an{@victoria, can you check the next two sentences? I just added them per Sida's comment.}Ideally, correct program samples shall be given higher probabilities thus we should observe higher percentage of programs being correct when it is closer to the top. To this end, we sort the prediction scores of the verifier, the generator and LEVER (as in \autoref{eq:join-reranking-prob}), and move the percentile threshold and measuring the percentage of correct programs in the top ranked programs. According to \autoref{fig:percentile_analysis}, the verifiers are generally better calibrated than the generators, especially when the threshold is in the lower percentiles. This indicates that it is easier for the verifiers to identify obvious mistakes in the programs with execution results as part of their input. Interestingly, when distinguishing between the top-ranked programs, the verifiers are poorly calibrated in three of the four tested datasets\footnote{Our hypothesis is that the programs ranked at the top have very similar form and execution results (\eg same type and range), making it hard for a small, though finetuned, model to discriminate.}.
However, the generators are generally better calibrated in this region, and combining the probability of the verifier and the generator yields the best results on all four benchmarks. 
More specifically, on the GSM8k dataset, where the calibration of both models are quite poor for top-ranking programs, their joint probability is surprisingly well-calibrated, showing that the two models complement each other on this dataset. 

\subsection{Quantitative Analysis}
\label{sec:quan-analysis}
\begin{figure}[t]
    \centering
    \begin{subfigure}[b]{0.47\textwidth}
         \centering
         \includegraphics[width=\linewidth]{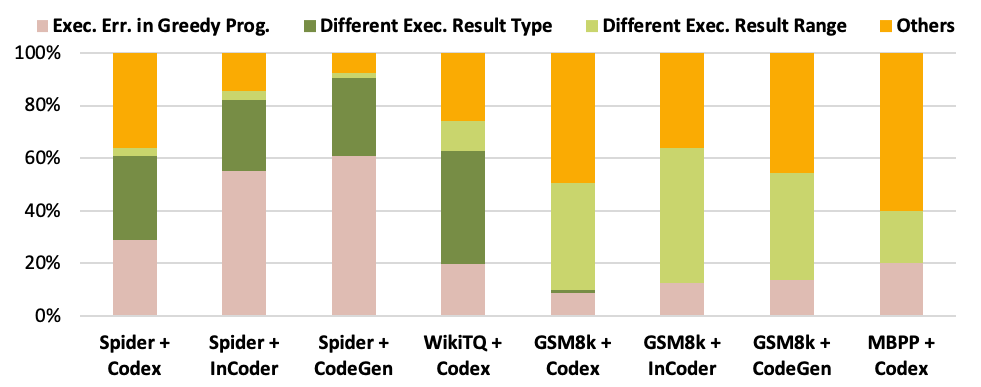}
         \caption{When \ours reranks a correct program at the top but the greedy decoding fails.}
         \label{fig:success-analysis}
    \end{subfigure}
    \newline
    \vspace{1pt}
    \newline
    \begin{subfigure}[b]{0.47\textwidth}
         \centering
         \includegraphics[width=\linewidth]{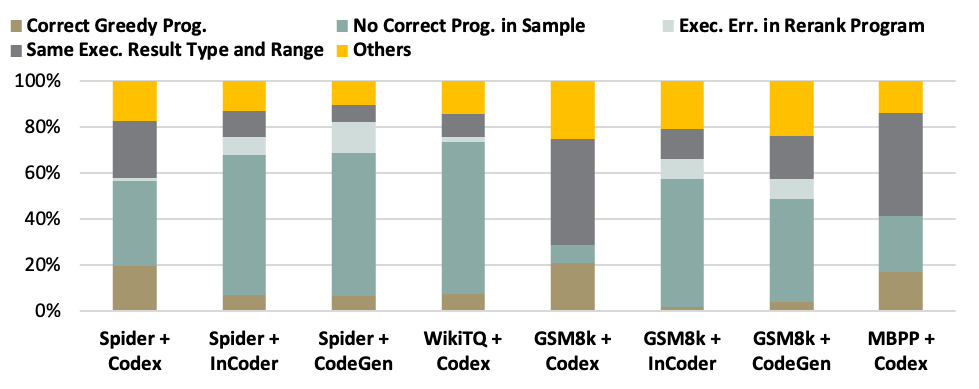}
         \caption{When \ours fails to rank a correct program at the top.}
         \label{fig:error-analysis}
     \end{subfigure}
     \caption{Quantitative analysis on when \ours succeeds and fails to improve code LLMs over greedy decoding.}
     \label{fig:quantitative-analysis}
\end{figure}
We present a quantitative analysis on why \ours successfully or failed to improve the performance of LLMs. 
According to \autoref{fig:quantitative-analysis}, when \ours reranks a program to replace another with higher generation probability, it is oftentimes because the execution results provide crucial information such as execution errors, variable type and range. This is consistent with our findings in \autoref{sec:main-ablation} about the importance of execution results for \ours.
It is also worth noticing that there are cases when \ours is still able to rerank the correct program when the error-free execution results are of the same type and range with the greedy program, \ie in ``others'' category. Our hypothesis is that this is when the program itself becomes the main feature for the verifiers to exploit. 
In addition, when \ours fails to rank correct programs to the top, the most common reason is that no correct program can be found in the samples (\ie upper-bound is reached), which is especially the case for weaker LMs. The second most common reason for \ours to fail is that the execution results of the incorrect program upon reranking has the same type and range as the correct program in the samples. In this case, execution results do not provide rich information for the verifiers thus \ours fails to improve code LLMs.

\section{Related Work}
\paragraph{Language-to-Code Generation.} 
Translating natural language to code is a long-standing challenge through all eras of artificial intelligence, including rule-based systems~\cite{DBLP:conf/afips/Woods73,DBLP:conf/anlp/TempletonB83}, structured prediction~\cite{DBLP:conf/aaai/ZelleM96,zettlemoyer2005,gulwani2014nlyze} and deep learning~\cite{xiao-etal-2016-sequence,DBLP:conf/acl/DongL16,rabinovich-etal-2017-abstract,DBLP:journals/corr/abs-1709-00103,Lin2017ProgramSF}.
Recently, pre-trained code language models~\cite{chen2021codex,wang2021codet5,fried2022incoder,nijkamp2022codegen,openai2022chatgpt} have demonstrated surprisingly strong performance in this problem across programming languages~\cite{lin2018nl2bash,yu-etal-2018-spider,austin2021mbpp,cobbe2021verifier,li2022alphacode}. A number of approaches were proposed to refine LLM sample selection, including test case execution~\cite{li2022alphacode}, cross-sample similarity~\cite{chen2021codex,li2022alphacode,shi2022mbr} and maximum mutual information~\cite{zhang2022coderreviewer} based filtering. Our work proposes a learnable verification module to judge the sample output of LLMs to further improve their performance.

\paragraph{Code Generation with Execution.} 
Previous code generation work have exploited execution results in different ways. Weakly-supervised learning approaches~\cite{berant2013freebase,pasupat-liang-2015-compositional,guu2017bridging} model programs as latent variables and use execution results to derive the supervision signal. Intermediate execution results were used to guide program search at both training~\cite{chen2018executionguided,DBLP:conf/nips/ChenST21} and inference time~\cite{wang2018execution}. When sampling at scale, majority voting based on the execution results has been shown effective for candidate selection~\cite{li2022alphacode,cobbe2021verifier}.~\citet{shi2022mbr} generalizes this principle by selecting samples that have the maximum concensus with other samples in the execution results. We propose to train a verification model to judge the correctness of code generation taking the execution results into account.

\paragraph{Learning to Verify.} Previous work have shown the effectiveness of learned verifiers for sample filtering in domains such as math QA~\cite{shen2021generateandrank,cobbe2021verifier} and commonsense QA~\cite{li2022diverse}, where the solution is mostly described in natural language. While it is more common to train the verifiers independently from the generator~\cite{cobbe2021verifier,li2022diverse},~\citet{shen2021generateandrank} jointly fine-tuned both at the same time. 
Previous work have also used different base LMs for the verifiers.
~\citet{cobbe2021verifier} uses GPT-3~\cite{brown2020language} while~\citet{li2022diverse} uses DeBERTa~\cite{he2020deberta}. 
Besides task-specific verifiers,~\citet{kadavath2022knowwhattheyknow} shows that large LMs can self-verify their output in a few-shot setting for a wide range of tasks.~\citet{chen2022codet} and other works~\cite{DBLP:journals/corr/abs-2009-05617,li2022alphacode} use LMs to generate test cases instead of directly judging the correctness of the output programs. In comparison, the setting of~\ours is closer to~\citet{li2022diverse} as we train the verifier separately and use a much smaller LM for it (approximately $0.5\%$ of the generator parameter size). 
We report the first set of comprehensive evaluation on language-to-code tasks, making use of the program execution results\footnote{While~\citet{kadavath2022knowwhattheyknow} also reported self-verification results on HumanEval, their approach does not leverage execution.}.
\paragraph{Discriminative Reranking.} Discriminative reranking approaches have long been used to further improve the performance of sequence generation tasks, including summarization~\cite{DBLP:journals/corr/WanCWLZ15}, machine translation~\cite{shen2004discriminative,lee2021discriminative}, dialogue response generation~\cite{DBLP:journals/corr/abs-1805-11752} and more rencently, code generation~\cite{yin-neubig-2019-reranking}. \ours can be viewed as a discriminative reranking framework.


\hide{
\paragraph{Reranking in NLP}
TODO
\paragraph{Large language models trained on code.}
TODO.
\paragraph{Execution-augmented code generation.}
TODO.
\paragraph{Semantic parsing.}
TODO.
\an{We should mention that though we are the first ones to try and made it work on NL2Code tasks, similar ideas are tried on other reasoning tasks, then mention verifier, consistency, etc.}
}

\section{Limitations}
In this work, we use execution information to verify the programs in \ours. However, the execution of the programs depends on at least one set of inputs (\eg arguments for a function) and adequate execution context (\eg databases), which may not be provided for certain applications. Moreover, we can not always assume that model-generated programs are safe to execute. In addition, we \patks{1} as the main evaluation metric in the experiments. While it is ideal for applications such as text-to-SQL and math reasoning where the users are only looking for answers to their questions, metrics as \patks{k} or \textsc{n@}\textit{k} could provide different perspectives for general programming tasks as MBPP.

\section{Conclusion}
We propose \ours, a simple approach for improving code LLMs on language-to-code tasks, by learning separate verification models to judge the correctness of the generated programs, taking their execution results into consideration. We show that it is possible to train verifiers approximately 0.5\% the size of the generators using supervised benchmark datasets. Instead of directly perform rejection sampling based on the verifier output, we show it is better to mix the generation and verfication probabilities for sample reranking.
\ours consistently improves the performance of code LLMs on four language-to-code tasks, and achieves new state-of-the-art results on all of them. Further analysis suggest that the program execution results are crucial for verification and the proposed approach is generalizable across different LLMs.

\section*{Acknowledgements}
The authors would like to thank Xi Ye, Tianyi Zhang, Mengzhou Xia, Luke Zettlemoyer, and the anonymous reviewers  for the useful discussion and comments.
\an{TODO}


\bibliography{main}
\bibliographystyle{icml2023}

\newpage
\appendix

\section{Additional Implementation Details}
\label{sec:more-imp-details}
\begin{table}[!h]
\small
\centering
    \begin{tabular}{p{0.18\linewidth}p{0.14\linewidth}p{0.14\linewidth}p{0.15\linewidth}p{0.14\linewidth}}
    \toprule
                        & \textbf{Spider}     & \textbf{WTQ}   & \textbf{GSM8k}    & \textbf{MBPP}\\\midrule
    \multicolumn{5}{c}{\textit{Verification Settings}} \\\midrule
    Input \quad Format  & NL+ SQL+ Exec. & NL+ SQL+ Exec. & NL+ Prog.+ Exec. & NL+ Prog.+ Exec. \\
    Normalize & \multirow{2}{*}{No} & \multirow{2}{*}{No} & \multirow{2}{*}{Yes} & \multirow{2}{*}{Yes} \\
    Gen. Prob. \\
    Batch Size$^*$ & 8       & 8       & 8       & 8 \\
    Downsample$^\dagger$ & 20     & 5       & 5       & 4 \\\midrule
    \multicolumn{5}{c}{\textit{\% of Positive Label}} \\\midrule
    Codex      & 68.0\%  & 47.6\%  & 61.1\%  & 56.6\%   \\
    InCoder    &  9.2\%  &  -      &  2.2\%  &  -       \\
    CodeGen    &  8.6\%  &  -      &  4.9\%  &  -       \\\midrule
    \multicolumn{5}{c}{\textit{\% of Unique Programs}} \\\midrule
    Codex      & 30.6\%  & 30.1\%  & 90.3\%  & 93.8\%   \\
    InCoder    & 94.2\%  &  -      & 99.2\%  &  -       \\
    CodeGen    & 94.5\%  &  -      & 99.3\%  &  -       \\
    
    \bottomrule
    \end{tabular}
    \caption{
    Detailed dataset-specific settings and statistics for the training of the verifiers. $^*$: number of examples per batch during training. $^\dagger$: number of program samples per example in the batch. The percentages are all among the sampled programs per example, and are all measured on the training set.
    }
    \label{tab:hyperparams}
\end{table}
\paragraph{Code LLM sampling.} We use temperature sampling to obtain program candidates given the input formats and sampling hyperparameters as described in \autoref{tab:hyperparams}. We set the temperature as $T=0.6$ for Codex and $T=0.8$ for InCoder and CodeGen, as the optimal temperatures for the best pass@k by referring to the original papers~\citep{fried2022incoder,nijkamp2022codegen}.
An ablation study on sampling budget is reported in \S\ref{sec:data-scaling}.

\paragraph{Few-shot prompt construction.} The numbers of few-shot exemplars to include in the prompt for different datasets are shown in \autoref{tab:datasets}. All exemplars are randomly sampled and ordered from the training set, with the exception of MBPP, where we use the 3 examples provided by the original dataset.
Full prompts used for each dataset are shown in \autoref{sec:prompts}.


\paragraph{Dataset-specific setups.} The detailed experiment setups for specific datasets are shown as \autoref{tab:hyperparams}. In particular, we use normalized probability for GSM8k and MBPP datasets as we find these two datasets can benefit from such normalization. We think this is because the Python programs have higher variance in length due to the flexible grammar and being more expressive. Moreover, the percentage of positive labels also denotes the ``random'' baseline, which is the expected execution accuracy by randomly picking from the sampled programs. This provides the additional perspective of the ability of the code LLMs, as well as the difficult of learning the verifiers for them.

\paragraph{Verifier Input Examples}
\label{sec:verifier-input-examples}
Here we show examples of the inputs to the verifiers for different datasets in \autoref{tab:verifier-inputs}.

\section{Additional Results}
\begin{table}[t]
\scriptsize
\centering
    \begin{subtable}[h]{0.5\textwidth}
        \centering
        \begin{tabular}{lc|ccc}
        \toprule
        \multicolumn{2}{l|}{\textbf{Target LLM}} & \multicolumn{3}{c}{\textbf{Source LLM}}         \\
        \multicolumn{2}{c|}{\textbf{\&}} & \multicolumn{3}{c}{\textbf{(\% Positive Labels)}}         \\\cline{3-5}
        \multicolumn{2}{c|}{\textbf{ML+EP}} & \textbf{Codex}       & \textbf{InCoder}      & \textbf{CodeGen}      \\
        \multicolumn{2}{c|}{\textbf{Baseline}} & \textbf{(64.0\%)} & \textbf{(9.2\%)} & \textbf{(8.6\%)} \\\midrule
        Codex             & 77.3          & \cellcolor{orange!50}82.0 (+4.7) & \cellcolor{orange!20}81.7 (+4.4)  & 80.8 (+3.5)  \\
        InCoder           & 41.2          & 46.4 (+5.2) & \cellcolor{orange!50}54.1 (+12.9) & \cellcolor{orange!20}47.6 (+6.4)  \\
        CodeGen           & 37.7          & 44.7 (+7.0) & \cellcolor{orange!20}48.9 (+11.2) & \cellcolor{orange!50}51.0 (+13.3) \\
        \bottomrule
        \end{tabular}
       \caption{Between the code LLMs transfer results on Spider. 
       }
       \label{tab:transfer-spider}
    \end{subtable}
    \newline
    \vspace{3pt}
    \newline
    \begin{subtable}[h]{0.5\textwidth}
        \centering
        \begin{tabular}{lc|ccc}
        \toprule
        \multicolumn{2}{l|}{\textbf{Target LLM}} & \multicolumn{3}{c}{\textbf{Source LLM}}         \\
        \multicolumn{2}{c|}{\textbf{\&}} & \multicolumn{3}{c}{\textbf{(\% Positive Labels)}}         \\\cline{3-5}
        \multicolumn{2}{c|}{\textbf{ML+EP}} & \textbf{Codex}       & \textbf{InCoder}      & \textbf{CodeGen}      \\
        \multicolumn{2}{c|}{\textbf{Baseline}} & \textbf{(53.4\%)} & \textbf{(2.3\%)} & \textbf{(5.0\%)} \\\midrule
        Codex             & \cellcolor{orange!20}72.1          & \cellcolor{orange!50}83.7 (+11.6) & 70.0 (-2.1)  & 71.9 (-0.2)  \\
        InCoder           & 4.3           & 8.3 (+4.0)   & \cellcolor{orange!20}11.9 (+7.6)  & \cellcolor{orange!50}12.3 (+8.0)  \\
        CodeGen           & 9.6           & 18.4 (+8.8)  & \cellcolor{orange!20}20.7 (+11.1) & \cellcolor{orange!50}22.1 (+12.5) \\
        \bottomrule
        \end{tabular}
        \caption{Between the code LLMs transfer results on GSM8k.}
        \label{tab:transfer-gsm}
     \end{subtable}
     \caption{Execution accuracy of training verifiers on the programs sampled from source code LLM and apply to the target code LLM. The \colorbox{orange!50}{best} and \colorbox{orange!20}{second best} performance per row is highlighted accordingly.
     }
     \vspace{-3pt}
     \label{tab:transfer}
\end{table}
\subsection{Transfer Learning between Code LLMs}
\label{sec:codelm-transfer-analysis}
One other way to avoid the cost of sampling from code LLMs is to train verifiers using samples from one LLM and directly apply to the programs sampled from a different LLM, \ie between LLM transfer, and we show the results of such on Spider and GSM8k in \autoref{tab:transfer}. From the results, we can first observe that \ours still non-trivially improves the baseline performance most of the time, with the exception of transferring from InCoder and CodeGen to Codex on the GSM8k dataset. This suggests that the knowledge learned by the verifiers are generalizable to different LLM outputs. Moreover, we can see that the transfer typically works better when the percentage of positive labels are closer, as the transfer is more successful between the InCoder and CodeGen models than that with Codex. 
These results show between-LLM transfer as an interesting way to reduce the training data need for \ours.

\subsection{Ablation on Base LMs for Verification}
\label{sec:base-model-ablation}
\begin{table}[]
\small
\centering
    \begin{tabular}{lcccc}
    \toprule
    \textbf{Base LMs}        & \textbf{Spider}     & \textbf{WTQ}   & \textbf{GSM8k}    & \textbf{MBPP}\\\midrule
    T5-base  & \textbf{82.0} & 64.8 & 82.4 & 76.8 \\
    T5-large & 81.9 & \textbf{65.0} & 82.5 & \textbf{77.3} \\
    T5-3B    & 83.1 & 64.7 & \textbf{84.4} & -    \\
    RoBERTa-large & - & 64.3 & \textbf{84.4} & -  \\
    \bottomrule
    \end{tabular}
    \caption{Ablations on using different base models for the verifiers. -: base LM not tested on this dataset.
    }
    \label{tab:base-model-ablation}
\end{table}
In this work, we treat the choice of base models for the verifiers as a hyperparameter and use the best performing model for further experiments. Here we show the performance of all the base models we attempted on the four datasets, with results in \autoref{tab:base-model-ablation}. 

\subsection{Training Example Scaling for WTQ and GSM8k}
\label{sec:train-examples-ablation-wtq-gsm}
\begin{figure}[t]
    \centering
         \includegraphics[width=0.95\linewidth]{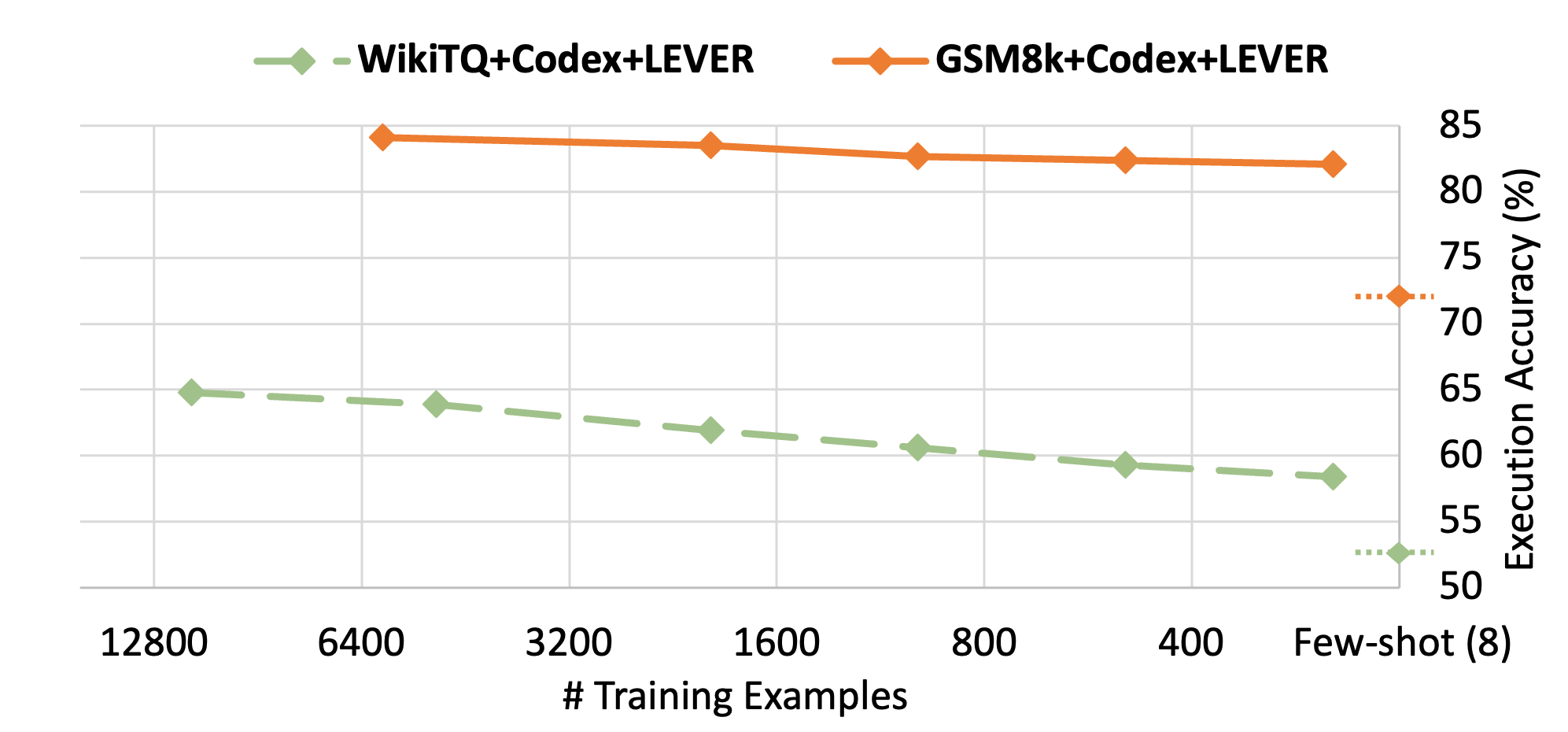}
         \caption{Ablation on number of training examples for Codex+\ours on the WTQ and GSM8k datasets. Data markers on the $y$-axis denote the ML+EP performances as baselines. T5-base is used for \ours.}
         \label{fig:train-examples-ablation-wtq-gsm}
\end{figure}
Due to space limit, we are only able to show the ablation in number of training example for Spider. Here in \autoref{fig:train-examples-ablation-wtq-gsm}, we show the results for WikiTQ and GSM8k as well. From the results, we can see that the learning of \ours is also very data efficient on those two benchmarks, as non-trivial improvements can be observed even when only 250 training examples are given.

\subsection{WikiTQ Results with the Official Evaluator}
\begin{table}[]
    \centering
    \small
    \begin{tabular}{lll}
        \toprule
        \textbf{Methods}     & \textbf{Dev}      & \textbf{Test}        \\\midrule
        \multicolumn{3}{c}{\textit{Previous Work without Finetuning}} \\
        Codex QA$^\ddag$ \citep{cheng2022binding}      & 49.3 & 47.6  \\
        Codex SQL$^\ddag$ \citep{cheng2022binding}     & 57.6 & 55.1  \\
        Codex Binder$^\ddag$ \citep{cheng2022binding}  & \textbf{62.6} & 61.9  \\\midrule
        \multicolumn{3}{c}{\textit{Previous Work with Finetuning}} \\
        TaPEX$^*$ \citep{liu2021tapex}   & 57.0 & 57.5  \\
        TaCube \citep{zhou2022tacube}    & 59.7 & 59.6  \\
        OmniTab \citep{jiang2022omnitab} & \;\;-    & \textbf{62.8}  \\\midrule
        \multicolumn{3}{c}{\textit{This Work Using code-davinci-002}} \\
        Greedy                 & 47.2             & 50.9          \\
        ML                     & 48.3             & 50.9          \\
        EP + ML                & 50.1             & 52.5          \\
        EP + Voting                 & 50.6             & 53.6          \\
        \oursf                 & 61.1$_{\pm0.2}$  & \textbf{62.9}$_{\pm0.2}$    \\
        \midrule
        Oracle                 & 70.9             & 74.6          \\
        \bottomrule
    \end{tabular}
    \caption{Execution accuracy on the WTQ dataset with the official WTQ executor. $^\ddag$: a normalizer to recognize date is added to the official executor.}
    \label{tab:codex-wtq-official}
\end{table}
Following \citet{cheng2022binding}, we fix the official evaluator of WikiTQ by normalizing units, Boolean values, etc. Here we also report the performance of \ours with previous work based on the official WikiTQ evaluator in \autoref{tab:codex-wtq-official}. From the results, we can see that \ours still presents the state-of-the-art result under this setting.

\begin{table*}[!h]
    \centering
    \footnotesize
    \begin{tabular}{p{0.95\linewidth}}
    \toprule
        \textsc{Spider/WikiTQ}: question + \bt{SQL} + \rt{linearized result table} \\
    \midrule
        \textbf{Input}: \\
        \texttt{-- question: List the name, born state and age of the heads of departments ordered by age.|}\\
        \texttt{\bt{-- SQL:|select name, born\_state, age from head join management on head.head\_id = management.head\_id order by age|}}\\
        \texttt{\rt{-- exec result:|/*| name born\_state age| Dudley Hart California 52.0| Jeff Maggert}} \\
        \texttt{\rt{Delaware 53.0|Franklin Langham Connecticut 67.0| Billy Mayfair California 69.0| }} \\
        \texttt{\rt{K. J. Choi Alabama 69.0|*/ }} \\
        \textbf{Output}: \texttt{no} \\
    \midrule
        \textsc{GSM8k}: question + \bt{idiomatic program} + \rt{answer variable}\\
    \midrule
        \textbf{Input}: \\
        \texttt{Carly recently graduated and is looking for work in a field she studied for. She sent 200 job applications to companies in her state, and twice that number to companies in other states. Calculate the total number of job applications she has sent so far. |}\\ 
        \texttt{\bt{n\_job\_apps\_in\_state = 200\newline n\_job\_apps\_out\_of\_state = n\_job\_apps\_in\_state * 2\newline answer = n\_job\_apps\_in\_state + n\_job\_apps\_out\_of\_state |}}\\
        \texttt{\rt{'answer': 600}} \\
        \textbf{Output}: \texttt{yes} \\
    \midrule
        \textsc{MBPP}: task description + \bt{function} + \rt{return type \& value} \\
    \midrule
        \textbf{Input}: \\
        \texttt{\# description} \\
        \texttt{Write a function to find the n-th power of individual elements in a list using lambda function.}\\\\
        \texttt{\bt{\# program}}\\
        \texttt{\bt{def nth\_nums(nums,n):}}\\
        \texttt{\bt{\quad\quad result\_list = list(map(lambda x: x ** n, nums))}}\\
        \texttt{\bt{\quad\quad return (result\_list)}}\\\\
        \texttt{\rt{\# execution}}\\
        \texttt{\rt{\# return: (list)=[1, 4, 9, 16, 25, 36, 49, 64, 81, 100]}}\\
        \texttt{\rt{\# return: (list)=[1000, 8000, 27000]}}\\
        \texttt{\rt{\# return: (list)=[248832, 759375]}}\\
        \textbf{Output}: \texttt{yes} \\
    \bottomrule
    \end{tabular}
    \caption{Examples of verifier inputs on the datasets. Some newlines are manually inserted for better display.
    }
    \label{tab:verifier-inputs}
\end{table*}

\subsection{Case Study}
Here we give some concrete examples to illustrate how \ours work and when does it fail in \autoref{tab:case-study-main}. In the first example from the Spider dataset, we can see that program candidate $\hy_2$ selects from the wrong table, which results in an execution error. This is easily detected by the verifier thus put a low verification probability on such program. Meanwhile, the execution result $\hz_1$ from program $\hy_1$ seems much more likely to be the answer of the question to the verifier. In the second example from WikiTQ, however, the execution results $\hz_1$ and $\hz_2$ do not provide clear information as they are both county names. In this case, the verifier does not possess much more meaningful information than the generator, thus not able to identify the incorrect program.
\begin{table*}[t]
    \centering
    \footnotesize
    \begin{tabular}{p{0.95\linewidth}}
    \toprule
        \textsc{Spider Example} \\
    \midrule
        \textbf{Question} $\bm{x}$: Find the total ranking points for each player and their first name.\\
        \textbf{Program} $\bm{\hy_1}$ \textbf{(\textcolor{teal}{correct})}: \\
        \hangindent=1em\texttt{select first\_name, sum(ranking\_points) from \hl{players} join rankings on player.player\_id = rankings.player\_id group by first\_name} \\
        \textbf{Program} $\bm{\hy_2}$ \textbf{(\textcolor{red}{incorrect})}: \\
        \hangindent=1em\texttt{select first\_name, sum(ranking\_points) from \hl{rankings} join players on rankings.player\_id = players.player\_id group by player\_id} \\
        \textbf{Execution info} $\bm{\hz_1}$: \\
        \texttt{Aastha | 68; Abbi | 304; Abbie | ...} \\
        \textbf{Execution info} $\bm{\hz_2}$: \\
        \texttt{ERROR: not column named ...} \\
    \midrule
        \textsc{WikiTQ Example} \\
    \midrule
        \textbf{Question} $\bm{x}$: When ranking the counties from first to last in terms of median family income, the first would be?\\
        \textbf{Program} $\bm{\hy_1}$ \textbf{(\textcolor{red}{incorrect})}: \\
        \hangindent=1em\texttt{select county from main\_table order by median\_family\_income\_number \hl{limit 1}} \\
        \textbf{Program} $\bm{\hy_2}$ \textbf{(\textcolor{teal}{correct})}: \\
        \hangindent=1em\texttt{select county from main\_table order by median\_family\_income\_number \hl{desc limit 1}} \\
        \textbf{Execution info} $\bm{\hz_1}$: \texttt{county | jefferson} \\
        \textbf{Execution info} $\bm{\hz_2}$: \texttt{county | sanders} \\
    \bottomrule
    \end{tabular}
    \caption{Case study for the WikiTQ and Spider datasets. Program $\hy_1$ is \textbf{ranked above} program $\hy_2$ in both examples. The main differences in the SQL programs that lead to error are \hl{highlighted}.\an{TODO: add GSM and MBPP example}\victoria{Also highlight execution info in the prompts since this is something unique about your verifier}\an{I did something in the next table, and it's probably better to switch the position of these two tables}}
    \label{tab:case-study-main}
\end{table*}

\section{Prompts for Few-shot Generation}
\label{sec:prompts}
Finally, we append the full prompts we used for few-shot prompting the code LLMs for Spider (\autoref{tab:spider-prompt}, \autoref{tab:spider-prompt-cont}, \autoref{tab:spider-prompt-cont-cont}), WikiTQ (\autoref{tab:wtq-prompt}, \autoref{tab:wtq-prompt-cont.}), GSM8k (\autoref{tab:gsm-prompt-1}, \autoref{tab:gsm-prompt-2}), and MBPP (\autoref{tab:mbpp-prompt}).

\begin{table*}[h]
    \centering
    \footnotesize
    \begin{tabular}{|p{0.95\linewidth}|}
    \toprule
\begin{lstlisting}[]
-- Translate natural language questions into SQL queries.

-- Example:

-- Database game_injury:
--  Table stadium: id, name, Home_Games, Average_Attendance, Total_Attendance, Capacity_Percentage
--  Table game: stadium_id, id, Season, Date, Home_team, Away_team, Score, Competition
--  Table injury_accident: game_id, id, Player, Injury, Number_of_matches, Source
-- Question: How many distinct kinds of injuries happened after season 2010?
-- SQL:
SELECT count(DISTINCT T1.Injury) FROM injury_accident AS T1 JOIN game AS T2 ON T1.game_id  =  T2.id WHERE T2.Season  >  2010

-- Example:

-- Database farm:
--  Table city: City_ID, Official_Name, Status, Area_km_2, Population, Census_Ranking
--  Table farm: Farm_ID, Year, Total_Horses, Working_Horses, Total_Cattle, Oxen, Bulls, Cows, Pigs, Sheep_and_Goats
--  Table farm_competition: Competition_ID, Year, Theme, Host_city_ID, Hosts
--  Table competition_record: Competition_ID, Farm_ID, Rank
-- Question: Return the hosts of competitions for which the theme is not Aliens?
-- SQL:
SELECT Hosts FROM farm_competition WHERE Theme !=  'Aliens'
\end{lstlisting}
\\\bottomrule
    \end{tabular}
    \caption{The prompt we use for the Spider dataset for few-shot generation with code LLMs. Only the first 2 exemplars are shown here, which is also the only two used for InCoder/CodeGen due to limits of model length and computation. 8 total exemplars are used for Codex, and the rest are shown in \autoref{tab:spider-prompt-cont} and \autoref{tab:spider-prompt-cont-cont}.
    }
    \label{tab:spider-prompt}
\end{table*}

\begin{table*}[h]
    \centering
    \footnotesize
    \begin{tabular}{|p{0.95\linewidth}|}
    \toprule
\begin{lstlisting}[]
-- Example:

-- Database school_finance:
--  Table School: School_id, School_name, Location, Mascot, Enrollment, IHSAA_Class, IHSAA_Football_Class, County
--  Table budget: School_id, Year, Budgeted, total_budget_percent_budgeted, Invested, total_budget_percent_invested, Budget_invested_percent
--  Table endowment: endowment_id, School_id, donator_name, amount
-- Question: Show the average, maximum, minimum enrollment of all schools.
-- SQL:
SELECT avg(Enrollment) ,  max(Enrollment) ,  min(Enrollment) FROM School

-- Example:

-- Database cre_Docs_and_Epenses:
--  Table Ref_Document_Types: Document_Type_Code, Document_Type_Name, Document_Type_Description
--  Table Ref_Budget_Codes: Budget_Type_Code, Budget_Type_Description
--  Table Projects: Project_ID, Project_Details
--  Table Documents: Document_ID, Document_Type_Code, Project_ID, Document_Date, Document_Name, Document_Description, Other_Details
--  Table Statements: Statement_ID, Statement_Details
--  Table Documents_with_Expenses: Document_ID, Budget_Type_Code, Document_Details
--  Table Accounts: Account_ID, Statement_ID, Account_Details
-- Question: Return the ids and details corresponding to projects for which there are more than two documents.
-- SQL:
SELECT T1.Project_ID ,  T1.Project_Details FROM Projects AS T1 JOIN Documents AS T2 ON T1.Project_ID  =  T2.Project_ID GROUP BY T1.Project_ID HAVING count(*)  >  2

-- Example:

-- Database local_govt_in_alabama:
--  Table Services: Service_ID, Service_Type_Code
--  Table Participants: Participant_ID, Participant_Type_Code, Participant_Details
--  Table Events: Event_ID, Service_ID, Event_Details
--  Table Participants_in_Events: Event_ID, Participant_ID
-- Question: List the type of the services in alphabetical order.
-- SQL:
SELECT Service_Type_Code FROM Services ORDER BY Service_Type_Code

-- Example:

-- Database cre_Theme_park:
--  Table Ref_Hotel_Star_Ratings: star_rating_code, star_rating_description
--  Table Locations: Location_ID, Location_Name, Address, Other_Details
--  Table Ref_Attraction_Types: Attraction_Type_Code, Attraction_Type_Description
--  Table Visitors: Tourist_ID, Tourist_Details
--  Table Features: Feature_ID, Feature_Details
--  Table Hotels: hotel_id, star_rating_code, pets_allowed_yn, price_range, other_hotel_details
--  Table Tourist_Attractions: Tourist_Attraction_ID, Attraction_Type_Code, Location_ID, How_to_Get_There, Name, Description, Opening_Hours, Other_Details
--  Table Street_Markets: Market_ID, Market_Details
--  Table Shops: Shop_ID, Shop_Details
--  Table Museums: Museum_ID, Museum_Details
--  Table Royal_Family: Royal_Family_ID, Royal_Family_Details
--  Table Theme_Parks: Theme_Park_ID, Theme_Park_Details
--  Table Visits: Visit_ID, Tourist_Attraction_ID, Tourist_ID, Visit_Date, Visit_Details
--  Table Photos: Photo_ID, Tourist_Attraction_ID, Name, Description, Filename, Other_Details
--  Table Staff: Staff_ID, Tourist_Attraction_ID, Name, Other_Details
--  Table Tourist_Attraction_Features: Tourist_Attraction_ID, Feature_ID
-- Question: Show the average price range of hotels that have 5 star ratings and allow pets.
-- SQL:
SELECT avg(price_range) FROM Hotels WHERE star_rating_code  =  "5" AND pets_allowed_yn  =  1
\end{lstlisting}
\\\bottomrule
    \end{tabular}
    \caption{The prompt we use for the Spider dataset for few-shot generation with code LLMs (Part 2), continued from \autoref{tab:spider-prompt}.
    }
    \label{tab:spider-prompt-cont}
\end{table*}

\begin{table*}[h]
    \centering
    \footnotesize
    \begin{tabular}{|p{0.95\linewidth}|}
    \toprule
\begin{lstlisting}[]
-- Example:

-- Database insurance_fnol:
--  Table Customers: Customer_ID, Customer_name
--  Table Services: Service_ID, Service_name
--  Table Available_Policies: Policy_ID, policy_type_code, Customer_Phone
--  Table Customers_Policies: Customer_ID, Policy_ID, Date_Opened, Date_Closed
--  Table First_Notification_of_Loss: FNOL_ID, Customer_ID, Policy_ID, Service_ID
--  Table Claims: Claim_ID, FNOL_ID, Effective_Date
--  Table Settlements: Settlement_ID, Claim_ID, Effective_Date, Settlement_Amount
-- Question: Find all the phone numbers.
-- SQL:
SELECT Customer_Phone FROM available_policies

-- Example:

-- Database cre_Theme_park:
--  Table Ref_Hotel_Star_Ratings: star_rating_code, star_rating_description
--  Table Locations: Location_ID, Location_Name, Address, Other_Details
--  Table Ref_Attraction_Types: Attraction_Type_Code, Attraction_Type_Description
--  Table Visitors: Tourist_ID, Tourist_Details
--  Table Features: Feature_ID, Feature_Details
--  Table Hotels: hotel_id, star_rating_code, pets_allowed_yn, price_range, other_hotel_details
--  Table Tourist_Attractions: Tourist_Attraction_ID, Attraction_Type_Code, Location_ID, How_to_Get_There, Name, Description, Opening_Hours, Other_Details
--  Table Street_Markets: Market_ID, Market_Details
--  Table Shops: Shop_ID, Shop_Details
--  Table Museums: Museum_ID, Museum_Details
--  Table Royal_Family: Royal_Family_ID, Royal_Family_Details
--  Table Theme_Parks: Theme_Park_ID, Theme_Park_Details
--  Table Visits: Visit_ID, Tourist_Attraction_ID, Tourist_ID, Visit_Date, Visit_Details
--  Table Photos: Photo_ID, Tourist_Attraction_ID, Name, Description, Filename, Other_Details
--  Table Staff: Staff_ID, Tourist_Attraction_ID, Name, Other_Details
--  Table Tourist_Attraction_Features: Tourist_Attraction_ID, Feature_ID
-- Question: Which transportation method is used the most often to get to tourist attractions?
-- SQL:
SELECT How_to_Get_There FROM Tourist_Attractions GROUP BY How_to_Get_There ORDER BY COUNT(*) DESC LIMIT 1
\end{lstlisting}
\\\bottomrule
    \end{tabular}
    \caption{The prompt we use for the Spider dataset for few-shot generation with code LLMs (Part 3), continued from \autoref{tab:spider-prompt} and \autoref{tab:spider-prompt-cont}.
    }
    \label{tab:spider-prompt-cont-cont}
\end{table*}
\begin{table*}[h]
    \centering
    \footnotesize
    \begin{tabular}{|p{0.95\linewidth}|}
    \toprule
\begin{lstlisting}[]
-- Translate natural language questions into SQL queries.

-- Example:

-- Database 204_126:
--  Table main_table: id (1), agg (0), place (t1), place_number (1.0), player (larry nelson), country (united states), score (70-72-73-72=287), score_result (287), score_number (70), score_number1 (70), score_number2 (72), score_number3 (73), score_number4 (72), to_par (-1), to_par_number (-1.0), money_lrb_rrb (playoff), money_lrb_rrb_number (58750.0)
-- Question: what was first place 's difference to par ?
-- SQL:
select to_par from main_table where place_number = 1

-- Example:

-- Database 204_522:
--  Table main_table: id (1), agg (0), boat_count (4911), boat_count_number (4911), boat_count_minimum (4951), boat_count_maximum (4955), name (ha-201), builder (sasebo naval arsenal), laid_down (01-03-1945), laid_down_number (1), laid_down_parsed (1945-01-03), laid_down_year (1945), laid_down_month (1), laid_down_day (3), launched (23-04-1945), launched_number (23), launched_parsed (1945-04-23), launched_year (1945), launched_month (4), launched_day (23), completed (31-05-1945), completed_number (31), completed_parsed (1945-05-31), completed_year (1945), completed_month (5), completed_day (31), fate (decommissioned 30-11-1945. scuttled off goto islands 01-04-1946)
-- Question: when was a boat launched immediately before ha-206 ?
-- SQL:
select name from main_table where launched_parsed < ( select launched_parsed from main_table where name = 'ha-206' ) order by launched_parsed desc limit 1

-- Example:

-- Database 204_877:
--  Table main_table: id (1), agg (0), place (1), place_number (1.0), position (mf), number (4), number_number (4.0), name (ryan hall), league_two (10), league_two_number (10.0), fa_cup (1), fa_cup_number (1.0), league_cup (0), league_cup_number (0.0), fl_trophy (3), fl_trophy_number (3.0), total (14), total_number (14.0)
-- Question: who scored more , grant or benyon ?
-- SQL:
select name from main_table where name in ( 'anthony grant' , 'elliot benyon' ) order by total_number desc limit 1

-- Example:

-- Database 204_400:
--  Table main_table: id (1), agg (0), district (1), district_number (1.0), senator (kenneth p. lavalle), party (republican), caucus (republican), first_elected (1976), first_elected_number (1976), counties_represented (suffolk), counties_represented_length (1)
--  Table t_counties_represented_list: m_id (1), counties_represented_list (suffolk)
-- Question: how many republicans were elected after 2000 ?
-- SQL:
select count ( * ) from main_table where party = 'republican' and first_elected_number > 2000
\end{lstlisting}
\\\bottomrule
    \end{tabular}
    \caption{The prompt we use for the WTQ dataset for few-shot generation with code LLMs (Part 1). 
    }
    \label{tab:wtq-prompt}
\end{table*}

\begin{table*}[h]
    \centering
    \footnotesize
    \begin{tabular}{|p{0.95\linewidth}|}
    \toprule
\begin{lstlisting}[]
-- Example:

-- Database 203_208:
--  Table main_table: id (1), agg (0), team (dinamo minsk), location (minsk), venue (dinamo, minsk), capacity (41040), capacity_number (41040.0), position_in_1993_94 (1), position_in_1993_94_number (1.0)
--  Table t_venue_address: m_id (1), venue_address (dinamo)
-- Question: what is the number of teams located in bobruisk ?
-- SQL:
select count ( team ) from main_table where location = 'bobruisk'

-- Example:

-- Database 203_60:
--  Table main_table: id (1), agg (0), outcome (winner), no (1), no_number (1.0), date (20 july 1981), date_number (20), date_parsed (1981-07-20), date_year (1981), date_month (7), date_day (20), championship (bastad, sweden), surface (clay), opponent_in_the_final (anders jarryd), score_in_the_final (6-2, 6-3), score_in_the_final_length (2)
--  Table t_championship_address: m_id (1), championship_address (bastad)
--  Table t_score_in_the_final_list: m_id (1), score_in_the_final_list (6-2)
--  Table t_score_in_the_final_list_first: m_id (1), score_in_the_final_list_first (6-2)
--  Table t_score_in_the_final_list_second: m_id (7), score_in_the_final_list_second (1-7)
--  Table t_score_in_the_final_list_first_number: m_id (1), score_in_the_final_list_first_number (6)
--  Table t_score_in_the_final_list_first_number1: m_id (1), score_in_the_final_list_first_number1 (6)
--  Table t_score_in_the_final_list_first_number2: m_id (1), score_in_the_final_list_first_number2 (2)
--  Table t_score_in_the_final_list_second_number: m_id (7), score_in_the_final_list_second_number (1)
--  Table t_score_in_the_final_list_second_number1: m_id (7), score_in_the_final_list_second_number1 (1)
--  Table t_score_in_the_final_list_second_number2: m_id (7), score_in_the_final_list_second_number2 (7)
-- Question: which month were the most championships played ?
-- SQL:
select date_month from main_table group by date_month order by count ( * ) desc limit 1

-- Example:

-- Database 203_462:
--  Table main_table: id (1), agg (0), year (2006), year_number (2006), division (4), division_number (4.0), league (usl pdl), regular_season (4th, heartland), regular_season_length (2), playoffs (did not qualify), open_cup (did not qualify)
--  Table t_regular_season_list: m_id (1), regular_season_list (4th)
-- Question: what year was more successful , 2012 or 2007 ?
-- SQL:
select year_number from main_table where year_number in ( 2012 , 2007 ) order by regular_season limit 1

-- Example:

-- Database 204_139:
-- Question: are their any other airports that are type `` military/public '' besides eagle farm airport ?
--  Table main_table: id (1), agg (0), community (antil plains), airport_name (antil plains aerodrome), type (military), coordinates (19 26'36''s 146 49'29''e/19.44333 s 146.82472 e)
-- SQL:
select ( select count ( airport_name ) from main_table where type = 'military/public' and airport_name != 'eagle farm airport' ) > 0
\end{lstlisting}
\\\bottomrule
    \end{tabular}
    \caption{The prompt we use for the WTQ dataset for few-shot generation with code LLMs (Part 2). 
    }
    \label{tab:wtq-prompt-cont.}
\end{table*}
\begin{table*}[h]
    \centering
    \footnotesize
    \begin{tabular}{|p{0.95\linewidth}|}
    \toprule
\begin{lstlisting}[]
## Cristina, John, Clarissa and Sarah want to give their mother a photo album for her birthday. Cristina brings 7 photos, John brings 10 photos and Sarah brings 9 photos. If the photo album has 40 slots available, how many photos does Clarissa need to bring in order to complete the photo album?
n_photo_cristina = 7
n_photo_john = 10
n_photo_sarah = 9
n_photo_total = n_photo_cristina + n_photo_john + n_photo_sarah
n_slots = 40
n_slots_left = n_slots - n_photo_total
answer = n_slots_left

## Katy, Wendi, and Carrie went to a bread-making party.  Katy brought three 5-pound bags of flour.  Wendi brought twice as much flour as Katy, but Carrie brought 5 pounds less than the amount of flour Wendi brought.  How much more flour, in ounces, did Carrie bring than Katy?
pound_flour_katy = 3 * 5
pound_flour_wendi = pound_flour_katy * 2
pound_flour_carrie = pound_flour_wendi - 5
pound_diff_carrie_katy = pound_flour_carrie - pound_flour_katy
ounce_diff_carrie_katy = pound_diff_carrie_katy * 16
answer = ounce_diff_carrie_katy

## James takes 2 Tylenol tablets that are 375 mg each, every 6 hours.  How many mg does he take a day?
mg_tylenol_per_tablet = 375
mg_tylenol_taken_each_time = 2 * mg_tylenol_per_tablet
hours_per_day = 24
times_per_day = hours_per_day / 6
mg_each_day = mg_tylenol_taken_each_time * times_per_day
answer = mg_each_day

## Kyle bakes 60 cookies and 32 brownies. Kyle eats 2 cookies and 2 brownies. Kyle's mom eats 1 cookie and 2 brownies. If Kyle sells a cookie for $1 and a brownie for $1.50, how much money will Kyle make if he sells all of his baked goods?
n_cookies = 60
n_brownies = 32
n_cookies_left_after_kyle = n_cookies - 2
n_brownies_left_after_kyle = n_brownies - 2
n_cookies_left_after_kyle_mom = n_cookies_left_after_kyle - 1
n_brownies_left_after_kyle_mom = n_brownies_left_after_kyle - 2
money_earned_kyle = n_cookies_left_after_kyle_mom * 1 + n_brownies_left_after_kyle_mom * 1.5
answer = money_earned_kyle

## There were 63 Easter eggs in the yard.  Hannah found twice as many as Helen.  How many Easter eggs did Hannah find?
n_easter_eggs = 63
unit_times = 2
total_units = unit_times + 1
n_easter_eggs_per_unit = n_easter_eggs / total_units
n_easter_eggs_helen = n_easter_eggs_per_unit * 1
n_easter_eggs_hannah = n_easter_eggs_per_unit * 2
answer = n_easter_eggs_hannah

## Ethan is reading a sci-fi book that has 360 pages. He read 40 pages on Saturday morning and another 10 pages at night. The next day he read twice the total pages as on Saturday. How many pages does he have left to read?
n_pages = 360
total_page_saturday = 40 + 10
total_page_next_day = total_page_saturday * 2
total_pages_read = total_page_saturday + total_page_next_day
n_pages_left = n_pages - total_pages_read
answer = n_pages_left
\end{lstlisting}
\\\bottomrule
    \end{tabular}
    \caption{The prompt we use for the GSM8k dataset for few-shot generation with code LLMs (Part 1). 
    }
    \label{tab:gsm-prompt-1}
\end{table*}

\begin{table*}[h]
    \centering
    \footnotesize
    \begin{tabular}{|p{0.95\linewidth}|}
    \toprule
\begin{lstlisting}[]
## A library has a number of books. 35% of them are intended for children and 104 of them are for adults. How many books are in the library?
percent_books_for_children = 0.35
percent_books_for_adults = 1.0 - percent_books_for_children
n_books_for_adults = 104
n_books_in_total = n_books_for_adults / percent_books_for_adults
answer = n_books_in_total

## Tyler has 21 CDs. He gives away a third of his CDs to his friend. Then he goes to the music store and buys 8 brand new CDs. How many CDs does Tyler have now?
n_cds_tyler = 21
percent_cds_given_away =  1.0 / 3.0
n_cds_left_after_giving_away = n_cds_tyler - n_cds_tyler * percent_cds_given_away
n_new_cds_purchased = 8
n_cds_now = n_cds_left_after_giving_away + n_new_cds_purchased
answer = n_cds_now
\end{lstlisting}
\\\bottomrule
    \end{tabular}
    \caption{The prompt we use for the GSM8k dataset for few-shot generation with code LLMs (Part 2).
    }
    \label{tab:gsm-prompt-2}
\end{table*}
\begin{table*}[h]
    \centering
    \footnotesize
    \begin{tabular}{|p{0.95\linewidth}|}
    \toprule
\begin{lstlisting}[]
# Write Python function to complete the task and pass the assertion tests. 

### Task Start ###
# These are the assertions for your function:
assert similar_elements((3, 4, 5, 6),(5, 7, 4, 10)) == (4, 5)

""" Write a function to find the similar elements from the given two tuple lists. """
def similar_elements(test_tup1, test_tup2):
    res = tuple(set(test_tup1) & set(test_tup2))
    return (res) 
### Task End ###

### Task Start ###
# These are the assertions for your function:
assert is_not_prime(2) == False

""" Write a python function to identify non-prime numbers. """
import math
def is_not_prime(n):
    result = False
    for i in range(2,int(math.sqrt(n)) + 1):
        if n % i == 0:
            result = True
    return result
### Task End ###

### Task Start ###
# These are the assertions for your function:
assert heap_queue_largest( [25, 35, 22, 85, 14, 65, 75, 22, 58],3)==[85, 75, 65] 

""" Write a function to find the largest integers from a given list of numbers using heap queue algorithm. """
import heapq as hq
def heap_queue_largest(nums,n):
    largest_nums = hq.nlargest(n, nums)
    return largest_nums
### Task End ###
\end{lstlisting}
\\\bottomrule
    \end{tabular}
    \caption{The prompt we use for the MBPP dataset for few-shot generation with code LLMs. 
    }
    \label{tab:mbpp-prompt}
\end{table*}





\end{document}